\documentclass[11pt]{article}

\usepackage[final]{acl}

\usepackage{times}
\usepackage{latexsym}
\usepackage{xcolor}
\usepackage{amsmath} 
\usepackage{amsfonts}
\usepackage{graphicx}
\usepackage{subcaption}
\usepackage{booktabs}
\usepackage{multirow}
\usepackage{booktabs}
\usepackage{array}
\usepackage{float}
\usepackage[table]{xcolor}
\newcommand{\best}[1]{\textbf{#1}}
\newcommand{\worst}[1]{\cellcolor{gray!18}#1}

\newcommand{\scoreDelta}[2]{#1\,{\scriptsize(#2)}}
\usepackage{url}
\usepackage{placeins}
\usepackage{tabularx}

\usepackage[T1]{fontenc}

\usepackage[utf8]{inputenc}

\usepackage{microtype}

\usepackage{inconsolata}

\usepackage{graphicx}

\title{Beyond Transfer Accuracy:\\Mechanism-Guided Controlled Adaptation for Low-Resource Languages}

\author{
 \textbf{Khumaisa Nur'aini\textsuperscript{1}},
 \textbf{Ayu Purwarianti\textsuperscript{2}},
 \textbf{Alham Fikri Aji\textsuperscript{1,3}},
 \textbf{Derry Wijaya\textsuperscript{1,4}}
\\
\\
 \textsuperscript{1}Monash University Indonesia,
 \textsuperscript{2}Institut Teknologi Bandung,
 \textsuperscript{3}MBZUAI,
 \textsuperscript{4}Boston University
\\
 \small{
\textbf{Correspondence:} 
\href{mailto:khumaisa.nuraini@monash.edu}{khumaisa.nuraini@monash.edu};
\href{mailto:derry.wijaya@monash.edu}{derry.wijaya@monash.edu}
}
}

\begin{document}
\maketitle
\begin{abstract}

Existing circuit discovery methods rely on templated tasks with clean counterfactuals, limiting their use on diverse natural text. We adapt Contextual Decomposition for Transformers (CD-T) for unstructured settings via label-balanced activation means and task-directional relevance scoring, enabling counterfactual-free circuit discovery. We leverage the discovered circuits for Circuit-Targeted Supervised Fine-Tuning (CT-SFT), restricting parameter updates to task-relevant heads and LayerNorm. Experiments on NusaX cross-lingual sentiment transfer show that CT-SFT is highly competitive for low-resource adaptation. While non-circuit sparse updates and full fine-tuning sometimes match target accuracy through capacity recruitment, CT-SFT most consistently avoids catastrophic forgetting, preserving source-language and related-task performance. Extensions to XNLI support the source-retention and intervention findings on a harder task and two model families, showing that circuit-targeted adaptation provides a more controlled, intervention-supported alternative to global fine-tuning.

\end{abstract}

\section{Introduction}

Mechanistic interpretability aims to reverse-engineer neural networks into human-understandable mechanisms by identifying internal components and circuits that implement model behavior~\cite{wang2023interpretability_in_the_wild_ioi, bereska2024mechanistic, olah2020zoom}. In transformer language models, circuit discovery has localized behavior to sparse sets of heads, neurons, or edges~\cite{wang2023interpretability_in_the_wild_ioi, conmy2023towards_automated_circuit_discovery, syed-etal-2024-attribution, hanna2024have_faith_in_faithfulness, hsu2025efficient_automated_circuit_discovery_cdt}. 

However, many circuit discovery methods are evaluated on structured or template-based tasks, where counterfactuals and task metrics are well-defined~\cite{wang2023interpretability_in_the_wild_ioi,conmy2023towards_automated_circuit_discovery,hsu2025efficient_automated_circuit_discovery_cdt}. Natural-text settings remain harder~\cite{sharkey2025open_problems_mi}, as diverse inputs make faithful counterfactuals expensive or ambiguous. This limitation is critical for low-resource language adaptation, where labeled data is scarce~\cite{joshi-etal-2020-state, aji-etal-2022-one, hu2020xtreme} and fine-tuning often risks instability or catastrophic forgetting~\cite{somayajula2024generalizable_stable_finetuning, du-etal-2023-measuring, mosbach-etal-2021-stability, dodge-etal-2020-finetuning}.

This challenge is especially visible in Indonesian local languages, where resources remain limited despite high linguistic diversity~\cite{aji-etal-2022-one,cahyawijaya-etal-2023-nusacrowd} and recent benchmarks show persistent gaps in model support~\cite{adilazuarda-etal-2025-nusaaksara,aji-cohn-2025-loraxbench}. NusaX~\cite{winata-etal-2023-nusax} provides a parallel sentiment benchmark for Indonesian and ten local languages, making it a useful testbed for low-resource source-to-target transfer. This setting also highlights a limitation of standard transfer evaluation: target-language gains may coincide with changes in source behavior, related tasks, broader classification ability, or perturbation consistency, which are connected to cross-lingual forgetting and behavioral testing~\cite{du-etal-2023-measuring,vu-etal-2022-overcoming,ribeiro2020checklist,wu2021polyjuice}.

To address this, we investigate whether circuits discovered from natural source-language data can guide controlled adaptation. We adapt Contextual Decomposition for Transformers (CD-T)~\cite{hsu2025efficient_automated_circuit_discovery_cdt} using label-balanced activation means and task-directional relevance scoring, enabling counterfactual-free discovery on natural text. We then introduce Circuit-Targeted Supervised Fine-Tuning (CT-SFT), which restricts parameter updates exclusively to task-relevant attention heads and LayerNorm. 

Our main evaluation uses NusaX sentiment transfer (Indonesian to Acehnese, Buginese, Javanese, and Minangkabau)~\cite{winata-etal-2023-nusax} with Llama-3.2-1B-Instruct~\cite{grattafiori2024llama3, meta2024llama321binstruct}, comparing CT-SFT against full fine-tuning and sparse non-circuit controls while evaluating retention on IndoNLU~\cite{wilie2020indonlu, saputri2018emotion} and SIB200~\cite{adelani2024sib200}; we further test scope using Llama-3.2-3B-Instruct for scalability and Llama-3.2-1B-Instruct with Qwen2.5-0.5B~\cite{qwen2, qwen2.5} on XNLI~\cite{conneau-etal-2018-xnli} for task and family breadth.

We ask four questions: 
(1) can counterfactual-free CD-T identify task-relevant circuit targets from natural text?
(2) can circuit-targeted tuning support low-resource transfer, and what do non-circuit sparse controls reveal about adaptation? 
(3) does circuit-targeted tuning provide more controlled adaptation beyond target accuracy, including source retention, related- and less-related task retention, perturbation consistency, and interpretability?
and (4) do the findings extend beyond the main NusaX Llama-1B setting, across model scale and a different multilingual classification task? 

\paragraph{Contributions.}
Our contributions are as follows. \\
\noindent\textbf{C1.} We adapt CD-T to natural-text classification using label-balanced activation means and task-directional relevance scoring, enabling counterfactual-free discovery of task-relevant circuit targets. \\
\noindent\textbf{C2.} We introduce circuit-targeted supervised fine-tuning (CT-SFT), a controlled low-resource adaptation strategy that updates only discovered task-relevant heads. \\
\noindent\textbf{C3.} We show that circuit tuning is competitive for transfer, while non-circuit sparse updates can also improve target accuracy, suggesting both mechanism editing and capacity recruitment. \\
\noindent\textbf{C4.} We evaluate adaptation beyond target accuracy using source retention, related- and less-related task retention, perturbation consistency, and faithfulness/knockout intervention metrics, showing that CT-SFT gives the most balanced and interpretable profile. \\
\noindent\textbf{C5.} We test the scope of CT-SFT with Llama-3.2-3B on NusaX and both Llama-1B and Qwen2.5-0.5B on XNLI, covering scalability and task/language-family breadth.

\section{Related Works}

Mechanistic interpretability (MI) localizes computations inside transformers~\citep{rai2024practical_review_mi, sharkey2025open_problems_mi}. Prior work has identified behavior-specific circuits using path patching ~\citep{wang2023interpretability_in_the_wild_ioi}, causal interventions, automated circuit discovery~\citep{conmy2023towards_automated_circuit_discovery}, and attribution-based methods~\citep{nanda2023attribution_patching, syed-etal-2024-attribution, hanna2024have_faith_in_faithfulness, zhang2025eapgp}. Contextual Decomposition for Transformers (CD-T) offers an alternative by decomposing activations into relevant and irrelevant streams~\citep{hsu2025efficient_automated_circuit_discovery_cdt}.

Many circuit discovery methods are evaluated on template-based tasks, where counterfactuals and reference states are easier to define than in unstructured natural text~\cite{sharkey2025open_problems_mi}. This gap is especially important for multilingual settings, as MI research has focused primarily on English~\citep{yu2025entangled}. Recent work has begun to examine multilingual interpretability and cross-lingual circuit structure, finding both shared pathways and language-specific specialization~\citep{resck-etal-2025-explainability, zhang2025same_but_different_multilingual}.

Recent work links mechanistic structure to adaptation: circuits can be reused across tasks~\citep{merullo2024circuit_component_reuse}, fine-tuning updates may concentrate in a small subset of heads~\citep{wang2025finetuning_mechanisms_circuit_analysis}, and circuit-guided tuning can update selected subgraphs/components instead of all parameters~\citep{li2025finetuning_subgraph_search,zhang2025exploring_translation_mechanism_llms}.

In contrast, we adapt CD-T~\cite{hsu2025efficient_automated_circuit_discovery_cdt} to unstructured, low-resource natural text and use the discovered circuits as targets for circuit-targeted supervised fine-tuning. This treats circuit discovery not only as a diagnosis but also as a controlled adaptation strategy for low-resource transfer with behavior preservation. By comparing circuit updates with non-circuit sparse controls, we analyze whether adaptation edits an existing task mechanism or recruits additional capacity.

\begin{figure*}[t]
    \centering
    \includegraphics[width=0.85\linewidth]{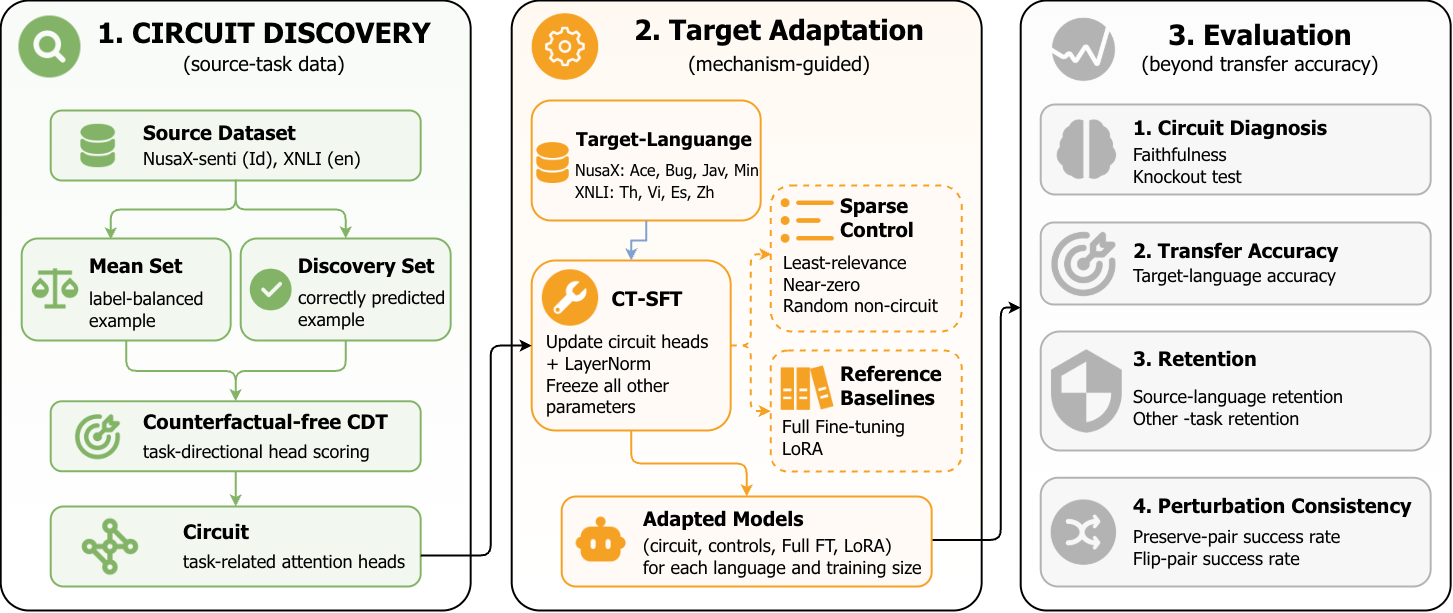}
    \caption{
    Overview of the controlled adaptation pipeline: counterfactual-free CD-T discovers a source-task circuit, CT-SFT adapts it to low-resource target languages, and evaluation measures circuit diagnosis, transfer, retention, and perturbation consistency.
    }
    \label{fig:full_pipeline}
\end{figure*}

\section{Background: Contextual Decomposition for Transformers (CD-T)}
\label{sec:background_cdt}

Contextual Decomposition for Transformers (CD-T)~\cite{hsu2025efficient_automated_circuit_discovery_cdt} identifies task-relevant components by decomposing each component activation into a baseline stream and an input-induced relevant stream. Let $a_x(s)$ be the activation of attention head $s$ on input $x$, and let $\mu(s)$ be the precomputed mean activation for that head. CD-T defines the baseline processing as $\gamma_s = \mu(s)$ and the input-induced deviation as $\beta_s = a_x(s) - \mu(s)$. In the original CD-T setting, $\mu$ is estimated from faithful counterfactual examples, such as name or year swaps in templated tasks like IOI, greater-than, and docstring completion. The $\gamma$ and $\beta$ streams are then propagated through the network using contextual decomposition rules to estimate the contribution $\beta_{s\to t}(x)$ from head $s$ to a target $t$.

CD-T ranks components using an unsigned relevance score based on the magnitude ratio between relevant and irrelevant streams:
\begin{equation}
\label{eq:cdt_original_score}
    R(s,t) = \frac{\lVert \beta_{s\to t} \rVert_1}{\lVert \gamma_{s\to t} \rVert_1}.
\end{equation}
This score favors heads whose relevant contribution is large relative to the baseline stream. Starting from the output readout, CD-T constructs a circuit by iteratively selecting the top-scoring heads and treating them as new targets for the next backward expansion step.

While effective for controlled settings, this formulation has two limitations for natural-text classification. First, natural text does not provide obvious faithful counterfactuals for estimating $\mu$. Second, the unsigned magnitude score can select heads with large but task-irrelevant or label-opposing contributions. The next section describes our counterfactual-free adaptation of CD-T for this setting.

\section{Counterfactual-Free Circuit Discovery for Natural Text}

Figure~\ref{fig:full_pipeline} summarizes our overall pipeline: we discover a source-language circuit using counterfactual-free CD-T, use the resulting circuit as the update target for CT-SFT, and evaluate the adapted models beyond target-transfer accuracy.

This section describes the discovery stage. We modify CD-T for natural-text classification in three steps. First, we replace counterfactual-based activation means with label-balanced means computed from natural source-language examples. Second, we rank heads by task-directional relevance, so that selected components are scored by whether they support the correct label rather than by magnitude alone. Third, we construct circuits through iterative backward expansion and apply score-based pruning to obtain a compact final circuit for adaptation.

\subsection{Label-Balanced Precomputed Mean}

A key dependency in CD-T is the precomputed mean $\mu$, which determines what is treated as background ($\gamma$) versus signal ($\beta$). In templated tasks, $\mu$ can be estimated from faithful counterfactuals, but such counterfactuals are difficult to construct for unstructured natural text. We therefore estimate $\mu$ from a label-balanced set of in-distribution examples $\mathcal{M}$, using equal numbers of examples per class to reduce class-directional leakage into the baseline, defining $\mu(s) = \mathbb{E}_{x \sim \mathcal{M}}[a_x(s)]$ to approximate a neutral operating point for each head.

\subsection{Task-Directional Relevance}

Original CD-T ranks components using an unsigned magnitude ratio. Still, large contributions are not necessarily task-supporting: a head may be non-discriminative or even anti-aligned with the gold label. To address this, we score heads by whether their relevant contribution supports the correct label direction.

At the logit readout, let $y_c$ be the correct label and $\mathcal{Y}_{\text{other}} = \mathcal{Y}\setminus\{y_c\}$. 
We define:
\begin{equation}
R(s,t) = \beta_{s\to t}(y_c) - \frac{1}{|\mathcal{Y}_{\text{other}}|} \sum_{y \in \mathcal{Y}_{\text{other}}} \beta_{s\to t}(y).
\end{equation}
For intermediate residual contributions, we define a task direction from the unembedding matrix:
\begin{equation}
    \mathbf{v}_{\text{task}} = W_U(y_c)-\frac{1}{|\mathcal{Y}_{\text{other}}|}\sum_{y \in \mathcal{Y}_{\text{other}}} W_U(y),
\end{equation}
and score the contribution by its projection onto this direction:
\begin{equation}
    R(s,t)= \frac{\beta_{s \to t}(x) \cdot \mathbf{v}_{\text{task}}}{\lVert \mathbf{v}_{\text{task}}\rVert_2}.
\end{equation}

\begin{figure}[ht]
    \centering
    \includegraphics[width=0.95\linewidth]{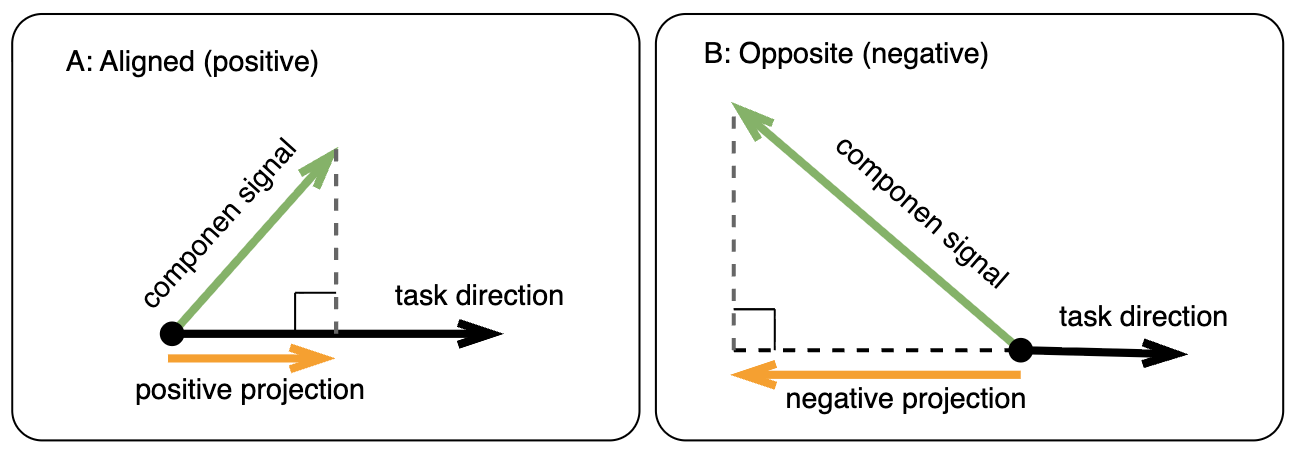}
    \caption{
    Illustration of task-directional relevance scoring. 
    Panel (A) shows an aligned component with a positive projection, while Panel (B) shows an opposing component with a negative projection.
    }
    \label{fig:directional_scoring}
\end{figure}

Our task-directional score is related to direct logit attribution (DLA) \cite{wang2023interpretability_in_the_wild_ioi,elhage2021mathematical}: both use the unembedding direction to estimate how an internal contribution supports the output decision. Unlike standard DLA, however, we apply this projection after CD-T decomposition, scoring the propagated relevant contribution \(\beta_{s \to t}(x)\) rather than the raw residual-stream activation. Thus, the score measures the task alignment of a head's input-induced contribution.
Positive scores indicate alignment with the correct-label direction, while negative scores indicate opposition. Figure~\ref{fig:directional_scoring} illustrates this distinction.

\subsection{Circuit Construction and Pruning}
\label{sec:circuit_construction_pruning}

Using the same iterative backward expansion procedure as in standard CD-T (Section~\ref{sec:background_cdt}), we construct our circuit by expanding backward from the output decision, guided by task-directional relevance.
These selected heads are then treated as new targets for the next depth, and the candidate circuit is obtained by aggregating selected heads across depths. 
We then apply score-based pruning by removing the selected heads with the lowest CD-T relevance scores (Appendix~\ref{app:training_discovery_protocol}), yielding a smaller final circuit $C$ for mechanism-guided tuning. 

\section{Mechanism-Guided Adaptation}

\subsection{Circuit-Targeted Supervised Fine-Tuning (CT-SFT) Setup}
\label{sec:ct_sft}

We use the discovered circuit as the update target for transfer: if the selected heads capture a task-relevant mechanism, then updating them should provide a more controlled route for adaptation than updating the full model. Given the final circuit $\mathcal{C}$ from Section~\ref{sec:circuit_construction_pruning}, Circuit-Targeted Supervised Fine-Tuning (CT-SFT) restricts optimization to heads in $\mathcal{C}$ and LayerNorm. We freeze MLPs, embeddings, the unembedding matrix, and all unselected heads, using head-level gradient masking to localize learning to the discovered circuit. This updates 2.355\% of Llama-1B and 2.410\% of Llama-3B parameters; details are in Appendix~\ref{app:update_budget}.

\subsection{Baselines and Sparse Controls}

\paragraph{Editing vs. recruitment.}
Table~\ref{tab:adaptation_methods} summarizes the adaptation methods. Circuit tuning tests \emph{mechanism editing}, since it updates heads identified as task-relevant. The sparse controls test whether similar low-resource transfer can also arise from \emph{capacity recruitment}, where non-circuit or weakly task-related heads are repurposed for the target task. Random NC is sampled from the complement of the discovered circuit, so it presupposes circuit discovery; it is therefore a sparse-capacity control, not a standalone random selection rule.

\begin{table}[ht]
\centering
\footnotesize
\setlength{\tabcolsep}{3pt}
\renewcommand{\arraystretch}{1.05}
\begin{tabular}{@{}l p{0.28\columnwidth} p{0.42\columnwidth}@{}}
\toprule
\textbf{Method} & \textbf{Heads} & \textbf{Role} \\
\midrule
Full FT & All params. & Global edit \\
Circuit & Circuit ($C$) & Mechanism-guided edit \\
Least-rel. & Negative-score & Negative recruitment \\
Near-zero & Near-zero-score & Indifferent recruitment \\
Random NC & Random heads outside $C$ & Non-circuit recruitment control \\
\bottomrule
\end{tabular}
\vspace{-0.7em}
\caption{Adaptation methods and mechanistic roles.}
\label{tab:adaptation_methods}
\end{table}

\section{Experimental Setup}

\subsection{Research Questions and Evaluation Map}

We evaluate our method through four research questions (RQs) covering intervention behavior~\citep{hanna2024have_faith_in_faithfulness,conmy2023towards_automated_circuit_discovery,chen2025rethinking_circuit_completeness}, transfer effectiveness, controlled adaptation beyond accuracy~\citep{du-etal-2023-measuring,ribeiro2020checklist,wu2021polyjuice}, and robustness across model scales and tasks.
Table~\ref{tab:rq_eval_map} maps these RQs to their corresponding evaluation protocols.
Full definitions for intervention metrics and datasets are provided in Appendices~\ref{app:faithfulness_knockout_metrics} and~\ref{app:dataset_details}.

\begin{table}[ht]
\centering
\footnotesize
\setlength{\tabcolsep}{2.5pt}
\renewcommand{\arraystretch}{1.05}
\begin{tabular}{
  >{\raggedright\arraybackslash}p{0.10\linewidth}
  >{\raggedright\arraybackslash}p{0.34\linewidth}
  >{\raggedright\arraybackslash}p{0.46\linewidth}
}
\toprule
\textbf{RQ} & \textbf{Evaluation} & \textbf{Purpose} \\
\midrule
RQ1 & Faithfulness, knockout & Identify task-relevant circuit targets \\
\addlinespace[1pt]
RQ2 & Transfer, sparse controls & Test transfer and recruitment \\
\addlinespace[1pt]
RQ3 & Retention, perturbations & Test controlled adaptation beyond accuracy \\
\addlinespace[1pt]
RQ4 & Llama-3B, XNLI & Test scale, task, and language-family robustness \\
\bottomrule
\end{tabular}
\vspace{-0.7em}
\caption{Research questions and evaluation protocols.}
\label{tab:rq_eval_map}
\end{table}

\subsection{Model}

We use Llama-3.2-1B-Instruct \citep{grattafiori2024llama3, meta2024llama321binstruct} as the main model for our reported experiments. To test scalability, we additionally run the NusaX experiment on Llama-3.2-3B-Instruct~\citep{grattafiori2024llama3, meta2024llama321binstruct}. For the XNLI~\cite{conneau-etal-2018-xnli} robustness experiment, we evaluate across two model families by using both Llama-3.2-1B-Instruct and Qwen2.5-0.5B~\cite{qwen2, qwen2.5}. Because these models perform near-chance on the three-way XNLI task before task adaptation, we first apply English-competence tuning before circuit discovery. Thus, the NusaX experiments use an instruction-tuned model with usable sentiment behavior out of the box, while the XNLI experiments first learn the English source task through competence tuning.

\begin{table*}[ht]
\centering
\footnotesize
\setlength{\tabcolsep}{3pt}
\renewcommand{\arraystretch}{1.05}
\begin{tabular}{
  >{\raggedright\arraybackslash}p{0.17\textwidth}
  >{\centering\arraybackslash}p{0.07\textwidth}
  >{\centering\arraybackslash}p{0.07\textwidth}
  >{\centering\arraybackslash}p{0.07\textwidth}
  >{\centering\arraybackslash}p{0.07\textwidth}
  >{\centering\arraybackslash}p{0.07\textwidth}
  >{\centering\arraybackslash}p{0.07\textwidth}
  >{\centering\arraybackslash}p{0.07\textwidth}
  >{\centering\arraybackslash}p{0.07\textwidth}
  >{\centering\arraybackslash}p{0.07\textwidth}
  >{\centering\arraybackslash}p{0.07\textwidth}
}
\toprule
\textbf{Method} 
& \multicolumn{5}{c}{\textbf{Faithfulness (\% full acc.)}} 
& \multicolumn{5}{c}{\textbf{Knockout (\% acc. drop)}} \\
\cmidrule(lr){2-6}
\cmidrule(lr){7-11}
& \textbf{Ind} & \textbf{Ace} & \textbf{Bug} & \textbf{Jav} & \textbf{Min}
& \textbf{Ind} & \textbf{Ace} & \textbf{Bug} & \textbf{Jav} & \textbf{Min} \\
\midrule

\multicolumn{11}{l}{\textit{Llama-3.2-1B-Instruct, avg. 62.8/512 heads after pruning, 2.355\% updated params}} \\
\addlinespace[1pt]
Circuit 
& \textbf{80.99} & \textbf{88.36} & \textbf{102.55} & \textbf{82.50} & \textbf{90.95}
& \textbf{52.82} & \textbf{37.93} & \textbf{17.35} & \textbf{43.33} & \textbf{34.91} \\

Least-relevant 
& 39.08 & 59.91 & 82.14 & 56.67 & 59.91
& 3.87 & 0.43 & -2.55 & 8.33 & -3.88 \\

Near-zero 
& 41.20 & 60.78 & 85.71 & 56.25 & 61.64
& 1.76 & 6.47 & 0.00 & 5.42 & 3.45 \\

Random NC 
& 42.96 & 60.78 & 87.76 & 56.67 & 61.64
& 14.08 & 17.67 & 1.53 & 20.83 & 15.52 \\

\midrule
\multicolumn{11}{l}{\textit{Llama-3.2-3B-Instruct, avg. 52.5/672 heads after pruning, 2.410\% updated params}} \\
\addlinespace[1pt]
Circuit 
& \textbf{73.44} & \textbf{101.63} & \textbf{99.44} & \textbf{89.50} & \textbf{90.69}
& \textbf{53.91} & \textbf{23.37} & \textbf{9.44} & \textbf{30.00} & \textbf{32.84} \\

Least-relevant 
& 34.77 & 68.48 & 80.56 & 60.00 & 61.76
& 31.25 & 6.52 & 4.44 & 13.50 & 13.73 \\

Near-zero 
& 37.50 & 70.11 & 80.56 & 61.50 & 60.78
& 9.77 & 1.63 & 3.89 & 6.50 & 7.35 \\

Random NC 
& 37.89 & 68.48 & 78.33 & 63.00 & 63.24
& 4.30 & -0.54 & 5.56 & 4.50 & 1.47 \\

\bottomrule
\end{tabular}
\vspace{-0.8em}
\caption{
Intervention evaluation for Llama-3.2-1B and 3B-Instruct. Scores use validation label-token accuracy averaged over four seeds. Pruned circuits contain 62.8/512 heads on average for 1B and 52.5/672 for 3B, corresponding to effective CT-SFT update budgets of 2.355\% and 2.410\% of the parameters updated by Full FT, respectively. Sparse controls are matched to the per-seed head budget. Faithfulness is retained full-model accuracy when only selected heads remain active; knockout is the accuracy drop when selected heads are mean-ablated. Faithfulness can exceed 100\% if ablating non-circuit heads removes noise.
}
\label{tab:faithfulness_knockout_by_model_lang}
\vspace{-1.5em}
\end{table*}

\subsection{Datasets and Evaluation Setting}
\label{sec:dataset_eval}

We utilize NusaX-Senti~\citep{winata-etal-2023-nusax} for low-resource transfer (Indonesian source to Acehnese, Buginese, Javanese, and Minangkabau targets). We evaluate target transfer and Indonesian source retention using 400 examples per language. Behavior preservation is assessed via Indonesian IndoNLU EmoT~\citep{wilie2020indonlu, saputri2018emotion} (related-task retention) and SIB200 topic classification~\citep{adelani2024sib200} (less-related retention), each using 400 examples. Perturbation consistency is evaluated using 200 label-preserving and 200 label-flipping pairs derived from NusaX-Senti, reported as pair-level success rates~\citep{ribeiro2020checklist}.

Finally, we evaluate robustness on XNLI~\citep{conneau-etal-2018-xnli}. To match the NusaX evaluation budget, we construct XNLI subsets containing 500 training, 100 validation, and 400 test examples per language. We use English for competence tuning and circuit discovery, with Thai, Vietnamese, Spanish, and Chinese as target languages. We report XNLI target transfer and English source retention. Unlike the Indonesian benchmarks, XNLI starts from an English NLI competence-tuned checkpoint. Therefore, IndoNLU EmoT, SIB200, and Indonesian perturbation evaluations are not directly comparable in this setting. Further dataset details and examples are provided in Appendix~\ref{app:dataset_details}.
\subsection{Training and Discovery Protocol}

Each target language is fine-tuned with $n \in \{25, 50, 75, 100\}$ examples across four seeds. Circuit discovery uses a label-balanced mean set and a correctly predicted discovery set from the source language. Heads are selected using task-directional relevance, followed by score-based pruning to obtain the final circuit. All sparse controls are then matched to the resulting post-pruning head budget for each seed.

The Llama-3.2-3B-Instruct NusaX experiments follow this same protocol as the main Llama-3.2-1B-Instruct evaluation. For XNLI, we first competence-tune Llama-3.2-1B-Instruct and Qwen2.5-0.5B on 250 English XNLI training examples, then perform circuit discovery on the competence-tuned English checkpoints. These circuits are then used for target-language CT-SFT. Full implementation, discovery, training, and effective update-budget details are provided in Appendices~\ref{app:model_compute_setup}--\ref{app:update_budget}.

\section{Results}

\subsection{RQ1: Counterfactual-free CD-T identifies task-relevant circuit targets in Llama}

\noindent\textbf{Key finding.} \textit{The discovered Llama circuits show higher faithfulness and stronger knockout effects than matched sparse controls.}

Table~\ref{tab:faithfulness_knockout_by_model_lang} reports validation-set intervention scores. Faithfulness measures retained label-token accuracy when only selected heads are preserved under mean ablation, while knockout measures the relative accuracy drop when selected heads are ablated. Across Indonesian and the target languages, Circuit achieves higher faithfulness and knockout than matched sparse controls, e.g., 53.91\% vs. 4.30\% knockout on Indonesian at 3B, indicating a stronger intervention target for CT-SFT. Sparse controls can retain non-zero faithfulness because transformer behavior may be distributed across redundant or self-repairing components~\citep{wang2023interpretability_in_the_wild_ioi,rushing2024explorations_self_repair}; however, their weaker knockout effects suggest they are less task-specific than the discovered circuit.

As a sensitivity check, we also evaluate zero ablation. This intervention is harsher and less aligned with CD-T's mean/reference decomposition, so keep-only faithfulness is more variable. Still, Circuit has the strongest average knockout for both Llama-1B and Llama-3B, showing that the selected heads remain load-bearing under a second ablation baseline. Since high zero-ablation faithfulness can occur without strong knockout, we treat knockout as the more diagnostic signal in this check. Detailed results are provided in Appendix~\ref{app:zero_ablation}.

\subsection{RQ2: Circuit tuning supports transfer, but sparse adaptation also involves recruitment}

\noindent\textbf{Key finding.} \textit{Circuit tuning is competitive for low-resource transfer, but least-relevant and random non-circuit updates can also improve target accuracy, suggesting that sparse adaptation may involve both mechanism editing and capacity recruitment.}

Table~\ref{tab:transfer_preservation_by_lang_llama1b} and Table~\ref{tab:reference_baselines} summarize target-language adaptation. Zero-shot results give the pre-adaptation baseline, while Full FT provides a high-plasticity reference that updates all parameters. Averaged across training sizes, Circuit is highly competitive among sparse methods; detailed trends are provided in Appendix~\ref{app:target_transfer_trends} and Figure~\ref{fig:target_transfer_trends}. However, non-circuit sparse updates can sometimes match or exceed Circuit, showing that transfer accuracy alone is insufficient to identify the most controlled or interpretable update strategy.

Because Random NC is sampled from outside the discovered circuit, it presupposes circuit discovery and should be interpreted as a sparse-capacity control rather than a standalone random selection rule. Its strength is consistent with sparse-update and lottery-ticket-style findings, where restricted trainable subsets can still support adaptation while reducing interference~\citep{frankle2019lottery,mehta2019sparse_transfer_lottery,ansell2022lottery_ticket_sparse_finetuning_acl}. More broadly, it is compatible with intervention-based self-repair observations, which suggest that transformer behavior can be distributed and supported by compensatory components rather than localized to a single non-redundant circuit~\citep{wang2023interpretability_in_the_wild_ioi,rushing2024explorations_self_repair}.

\subsection{RQ3: Circuit tuning provides more controlled adaptation beyond target accuracy}

\noindent\textbf{Key finding.} \textit{Non-circuit sparse updates can be competitive on individual metrics, but Circuit provides the most balanced and interpretable profile across faithfulness, transfer, retention, and perturbation consistency.}

We use the preservation evaluations defined in Section~\ref{sec:dataset_eval} to test whether sparse transfer gains remain stable beyond target accuracy.

\begin{table*}[t]
\centering
\footnotesize
\setlength{\tabcolsep}{3.5pt}
\renewcommand{\arraystretch}{1.08}
\begin{tabular*}{\textwidth}{@{\extracolsep{\fill}}llcccccc@{}}
\toprule
\textbf{Method} & \textbf{Evaluation}
& \textbf{Zero-shot}
& \textbf{Ace} & \textbf{Bug} & \textbf{Jav} & \textbf{Min} & \textbf{Avg.} \\
\midrule

Zero-shot
& Transfer
& --
& 52.25 & 39.00 & 56.00 & 52.00 & 49.81 \\

\midrule

Circuit 
& Transfer & --
& \best{50.00} & 42.90 & \best{63.50} & 61.30 & \best{54.40} \\
& Source & 73.50
& 69.20 & 68.40 & 70.30 & \best{70.30} & 69.50 \\
& Emotion & 41.25
& \best{40.10} & 39.10 & \best{41.40} & 40.00 & \best{40.20} \\
& Topic & 51.25
& \best{68.40} & \best{69.40} & \best{69.60} & 69.60 & \best{69.30} \\
& Pres. Pair & 64.00
& 64.30 & 63.80 & 65.70 & 65.00 & 64.70 \\
& Flip Pair & 81.00
& 79.90 & 78.40 & 79.90 & 79.70 & 79.50 \\

\midrule

Least-relevant
& Transfer & --
& \scoreDelta{48.20}{-1.80} 
& \scoreDelta{42.60}{-0.30} 
& \scoreDelta{60.50}{-3.00} 
& \best{\scoreDelta{61.50}{+0.20}} 
& \scoreDelta{53.20}{-1.20} \\
& Source & 73.50
& \scoreDelta{70.20}{+1.00} 
& \scoreDelta{67.80}{-0.60} 
& \best{\scoreDelta{71.10}{+0.80}} 
& \scoreDelta{69.70}{-0.60} 
& \scoreDelta{69.70}{+0.20} \\
& Emotion & 41.25
& \worst{\scoreDelta{32.80}{-7.30}} 
& \worst{\scoreDelta{33.00}{-6.10}} 
& \worst{\scoreDelta{32.90}{-8.50}} 
& \worst{\scoreDelta{32.70}{-7.30}} 
& \worst{\scoreDelta{32.90}{-7.30}} \\
& Topic & 51.25
& \best{\scoreDelta{68.40}{0.00}} 
& \scoreDelta{69.10}{-0.30} 
& \scoreDelta{69.10}{-0.50} 
& \best{\scoreDelta{70.20}{+0.60}} 
& \scoreDelta{69.20}{-0.10} \\
& Pres. Pair & 64.00
& \scoreDelta{64.20}{-0.10} 
& \scoreDelta{62.60}{-1.20} 
& \best{\scoreDelta{65.80}{+0.10}} 
& \scoreDelta{64.00}{-1.00} 
& \scoreDelta{64.10}{-0.60} \\
& Flip Pair & 81.00
& \worst{\scoreDelta{75.50}{-4.40}} 
& \worst{\scoreDelta{67.70}{-10.70}} 
& \worst{\scoreDelta{75.10}{-4.80}} 
& \worst{\scoreDelta{76.80}{-2.90}} 
& \worst{\scoreDelta{73.80}{-5.70}} \\

\midrule

Near-zero
& Transfer & --
& \worst{\scoreDelta{47.10}{-2.90}} 
& \worst{\scoreDelta{39.90}{-3.00}} 
& \worst{\scoreDelta{59.10}{-4.40}} 
& \worst{\scoreDelta{55.80}{-5.50}} 
& \worst{\scoreDelta{50.50}{-3.90}} \\
& Source & 73.50
& \worst{\scoreDelta{66.20}{-3.00}} 
& \worst{\scoreDelta{65.50}{-2.90}} 
& \worst{\scoreDelta{65.90}{-4.40}} 
& \worst{\scoreDelta{64.40}{-5.90}} 
& \worst{\scoreDelta{65.50}{-4.00}} \\
& Emotion & 41.25
& \scoreDelta{37.50}{-2.60} 
& \best{\scoreDelta{40.40}{+1.30}} 
& \scoreDelta{39.60}{-1.80} 
& \best{\scoreDelta{40.10}{+0.10}} 
& \scoreDelta{39.40}{-0.80} \\
& Topic & 51.25
& \worst{\scoreDelta{57.20}{-11.20}} 
& \worst{\scoreDelta{59.80}{-9.60}} 
& \worst{\scoreDelta{59.40}{-10.20}} 
& \worst{\scoreDelta{58.50}{-11.10}} 
& \worst{\scoreDelta{58.70}{-10.60}} \\
& Pres. Pair & 64.00
& \worst{\scoreDelta{62.30}{-2.00}} 
& \worst{\scoreDelta{62.10}{-1.70}} 
& \worst{\scoreDelta{62.20}{-3.50}} 
& \worst{\scoreDelta{61.30}{-3.70}} 
& \worst{\scoreDelta{62.00}{-2.70}} \\
& Flip Pair & 81.00
& \best{\scoreDelta{83.00}{+3.10}} 
& \best{\scoreDelta{82.80}{+4.40}} 
& \best{\scoreDelta{81.30}{+1.40}} 
& \best{\scoreDelta{80.50}{+0.80}} 
& \best{\scoreDelta{81.90}{+2.40}} \\

\midrule

Random NC
& Transfer & --
& \scoreDelta{48.00}{-2.00} 
& \best{\scoreDelta{44.50}{+1.60}} 
& \scoreDelta{61.80}{-1.70} 
& \scoreDelta{60.30}{-1.00} 
& \scoreDelta{53.60}{-0.80} \\
& Source & 73.50
& \best{\scoreDelta{70.30}{+1.10}} 
& \best{\scoreDelta{69.70}{+1.30}} 
& \scoreDelta{70.90}{+0.60} 
& \scoreDelta{70.10}{-0.20} 
& \best{\scoreDelta{70.30}{+0.80}} \\
& Emotion & 41.25
& \scoreDelta{35.80}{-4.30} 
& \scoreDelta{35.10}{-4.00} 
& \scoreDelta{37.10}{-4.30} 
& \scoreDelta{36.30}{-3.70} 
& \scoreDelta{36.10}{-4.10} \\
& Topic & 51.25
& \scoreDelta{63.30}{-5.10} 
& \scoreDelta{66.10}{-3.30} 
& \scoreDelta{63.80}{-5.80} 
& \scoreDelta{64.00}{-5.60} 
& \scoreDelta{64.30}{-5.00} \\
& Pres. Pair & 64.00
& \best{\scoreDelta{65.00}{+0.70}} 
& \best{\scoreDelta{64.40}{+0.60}} 
& \scoreDelta{65.20}{-0.50} 
& \best{\scoreDelta{65.20}{+0.20}} 
& \best{\scoreDelta{64.90}{+0.20}} \\
& Flip Pair & 81.00
& \scoreDelta{80.80}{+0.90} 
& \scoreDelta{79.30}{+0.90} 
& \scoreDelta{78.10}{-1.80} 
& \scoreDelta{78.90}{-0.80} 
& \scoreDelta{79.30}{-0.20} \\

\bottomrule
\end{tabular*}
\vspace{-0.8em}
\caption{
Transfer and preservation of behavior after target-language adaptation. Scores are percentages averaged across training sizes and seeds; parentheses show differences from Circuit. Bold marks the best score, gray shading the lowest. Random NC denotes random non-circuit updates.
Full FT is reported separately in Table~\ref{tab:reference_baselines}; detailed training-size trends are shown in Appendix~\ref{app:target_transfer_trends}.
}
\label{tab:transfer_preservation_by_lang_llama1b}
\vspace{-1.5em}
\end{table*}

\paragraph{Source-language retention.}
On Indonesian source-task retention, Random non-circuit achieves the strongest average score (Table~\ref{tab:transfer_preservation_by_lang_llama1b}). One possible explanation is that Random non-circuit does not directly update the discovered source-task circuit, thereby leaving the original sentiment mechanism less disrupted. At the same time, the updated non-circuit heads can provide additional capacity for target-language adaptation. This supports the recruitment interpretation: non-circuit updates can preserve the original circuit while repurposing other heads to support transfer.

\paragraph{Other-task retention.}
The circuit is more stable than non-circuit updates on the two other-task retention evaluations (Table~\ref{tab:transfer_preservation_by_lang_llama1b}). This suggests that localizing updates to the discovered circuit does not strongly disrupt the broader classification logic used by related and less-related tasks. In contrast, non-circuit updates can support transfer through recruitment, but the recruited heads are not guaranteed to preserve other task-relevant computations. Since Least-relevant and Near-zero are defined only with respect to sentiment relevance, updating them may still disturb affective or topic-classification behavior, which may explain the larger drops on IndoNLU EmoT and SIB200.

\paragraph{Perturbation consistency.}
Perturbation results show that label-preserving and label-flipping consistency capture different behaviors (Table~\ref{tab:transfer_preservation_by_lang_llama1b}). Random non-circuit is slightly better than Circuit on label-preserving pairs, while Near-zero obtains the highest label-flipping score. This suggests that Near-zero heads are more plastic: because they are weakly tied to the original sentiment circuit, they can be repurposed more easily when explicit sentiment cues are flipped. However, this flexibility is not equivalent to a better sentiment mechanism, since Near-zero is also weakest on transfer, source retention, topic retention, and label-preserving perturbations. Thus, Near-zero is more responsive to label-changing edits, while Circuit provides a more balanced profile across transfer, retention, and consistency.

\paragraph{Reference baselines.}
Table~\ref{tab:reference_baselines} reports Full FT and LoRA as same-data/same-protocol reference baselines, not exhaustively tuned optima. All methods use the same target examples, seeds, training setup, and evaluation pipeline, so the comparison isolates behavior under a shared protocol rather than proving dominance over all possible FT or PEFT variants. Full FT slightly improves average transfer over Circuit but substantially weakens preservation. LoRA preserves source retention, emotion retention, and label-flipping consistency well, but is weaker on target transfer, SIB200 Topic retention, and label-preserving perturbations. Thus, LoRA is effective for sentiment-adjacent preservation, while Circuit provides a stronger broader controlled-adaptation profile. A per-language Circuit--LoRA breakdown is provided in Appendix~\ref{app:lora_circuit_details}.

\begin{table}[ht]
\centering
\footnotesize
\setlength{\tabcolsep}{3pt}
\renewcommand{\arraystretch}{1.05}
\begin{tabular}{lccccc}
\toprule
\textbf{Eval} 
& \textbf{Full FT} 
& \textbf{LoRA} 
& \textbf{Circuit} 
& $\Delta_{\text{FT}}$ 
& $\Delta_{\text{LoRA}}$ \\
\midrule
Trans.     
& 55.30
& 50.32
& 54.41
& +0.89 & -4.09 \\

Source        
& 62.43
& 72.98
& 69.55
& -7.11 & +3.44 \\

Emot.       
& 6.09
& 41.07
& 40.16
& -34.07 & +0.92 \\

Topic         
& 12.17
& 48.51
& 69.29
& -57.12 & -20.77 \\

Pres.
& 60.09
& 63.04
& 64.72
& -4.62 & -1.68 \\

Flip     
& 47.98
& 82.54
& 79.47
& -31.48 & +3.07 \\
\bottomrule
\end{tabular}
\vspace{-0.6em}
\caption{
NusaX reference baselines. Scores are averaged over seed-level averages; deltas are relative to Circuit.
}
\label{tab:reference_baselines}
\vspace{-1em}
\end{table}

\paragraph{LayerNorm and localization.}
LN-only remains close to the pre-adaptation baseline, especially on transfer and Topic retention (Table~\ref{tab:layernorm_ablation}). In contrast, heads-only no-LN recovers most of CT-SFT's transfer and Topic gains. This indicates that LayerNorm alone is not a competitive adaptation path and cannot explain CT-SFT's main gains. The selected heads carry the primary adaptive capacity, while LayerNorm provides an auxiliary, stabilizing, or complementary update. Detailed per-language LayerNorm results and heads-only sparse-control comparisons are provided in Appendix~\ref{app:layernorm_details}.

\begin{table}[ht]
\centering
\small
\setlength{\tabcolsep}{2.5pt}
\renewcommand{\arraystretch}{1.05}
\begin{tabular}{@{}lrrrr@{}}
\toprule
\textbf{Eval.} 
& \textbf{Pre} 
& \textbf{LN} 
& \textbf{Heads} 
& \textbf{Heads+LN} \\
\midrule
Transfer   & 49.81 & 50.70 & 54.22 & \best{54.41} \\
Source     & 73.50 & \best{73.84} & 71.94 & 69.55 \\
Emotion    & 41.25 & 40.98 & \best{41.30} & 40.16 \\
Topic      & 51.25 & 51.71 & 67.48 & \best{69.29} \\
Pres. Pair & 64.00 & 63.74 & 64.10 & \best{64.72} \\
Flip Pair  & 81.00 & 82.66 & \best{85.81} & 79.47 \\
\bottomrule
\end{tabular}
\vspace{-0.6em}
\caption{
Aggregate Llama-1B LayerNorm ablation results, averaged over training sizes and seeds. Pre is the pre-adaptation model; LN updates only LayerNorm/RMSNorm; Heads updates selected attention heads; Heads+LN is CT-SFT. Higher is better.
}
\label{tab:layernorm_ablation}
\end{table}
 
\subsection{RQ4: Does the finding extend beyond the main Llama-1B NusaX setting?}

\noindent\textbf{Key finding.} \textit{The core findings extend beyond the main Llama-1B NusaX setting: Circuit remains load-bearing under intervention and avoids the severe forgetting of Full FT, while non-circuit transfer gains recur, supporting the recruitment pattern from RQ2.}

\paragraph{Scalability on Llama-3.2-3B-Instruct.}
Tables~\ref{tab:faithfulness_knockout_by_model_lang} and~\ref{tab:llama3b_scalability_avg} test whether the NusaX findings hold at 3B scale. The intervention results strengthen RQ1: Circuit achieves much higher faithfulness and knockout than matched controls, while Random NC has near-zero knockout. Thus, the discovered heads remain strong intervention targets in the larger model.

\begin{table}[ht]
\centering
\footnotesize
\setlength{\tabcolsep}{2.5pt}
\renewcommand{\arraystretch}{1.06}
\begin{tabular*}{\columnwidth}{@{\extracolsep{\fill}}lcccc@{}}
\toprule
\textbf{Method} 
& \textbf{Transfer} 
& \textbf{Source} 
& \textbf{Emotion} 
& \textbf{Topic} \\
\midrule
Zero-shot
& 30.12 & 56.00 & 34.50 & 20.00 \\

\midrule
Circuit
& 58.68
& 76.99
& 36.84
& 68.54 \\

Least-rel.
& 59.66
& 76.70
& 34.91
& 58.11 \\

Near-zero
& 58.01
& 76.25
& 30.88
& \worst{45.34} \\

Rand. NC
& 58.70
& \best{77.11}
& 33.51
& \best{70.30} \\

\midrule
Full FT
& \best{60.08}
& 71.45
& \worst{1.01}
& \worst{4.99} \\

LoRA
& 55.14
& 73.05
& \best{38.16}
& 32.39 \\

\bottomrule
\end{tabular*}
\vspace{-0.6em}
\caption{
Llama-3B NusaX results. Scores are averaged over four seeds after averaging across target languages and training sizes. Full per-language sparse-control and reference-baseline results are provided in Appendices~\ref{app:llama3b_sparse_controls} and~\ref{app:llama3b_reference_baselines}.
}
\label{tab:llama3b_scalability_avg}
\vspace{-1em}
\end{table}

The downstream results also support RQ3. Full FT gives the highest transfer but severely damages related and broader behavior, reducing Emotion and Topic retention to near-zero levels. LoRA is less destructive, but remains weak on Topic retention and target transfer. Circuit therefore gives a stronger controlled-adaptation profile: it remains competitive in transfer while avoiding the severe behavioral collapse of Full FT and the narrower Topic-retention profile of LoRA.

The sparse controls reinforce the recruitment pattern from RQ2. Random NC matches Circuit on transfer and source retention and slightly exceeds it on Topic retention, showing that non-circuit heads can sometimes provide useful adaptation capacity. However, Random NC has near-zero knockout and is not selected by source-task intervention evidence. Thus, circuit targeting is valuable not because it always dominates every sparse control, but because it gives a comparable profile from a principled, interpretable, knockout-supported update location. The 3B results also show that sparse controls are not interchangeable: relative to Circuit, Near-zero loses 23.2 points, and Least-relevant loses 10.4 points on SIB200 Topic retention.

\begin{table}[ht]
\centering
\small
\setlength{\tabcolsep}{4pt}
\renewcommand{\arraystretch}{1.02}
\begin{tabular}{@{}lrrrr@{}}
\toprule
\textbf{Model} 
& \textbf{Zero} 
& \textbf{Circuit} 
& \textbf{LoRA} 
& \textbf{Full FT} \\
\midrule
Llama-1B & 14.16 & 14.10 & 14.08 & 135.74 \\
Llama-3B & 9.60  & 9.55  & 9.54  & 16.10 \\
\bottomrule
\end{tabular}
\vspace{-0.6em}
\caption{
Indonesian Wikipedia perplexity after adaptation.
Lower is better. Scores are averaged over target-language adaptation settings.
}
\label{tab:ppl_diagnostic}
\end{table}

As a diagnostic for broad language-model drift, we evaluate Indonesian Wikipedia perplexity after adaptation (Table~\ref{tab:ppl_diagnostic}). LoRA remains close to Circuit in perplexity, so its weaker SIB200 Topic retention is not simply explained by broad language-model drift. Full FT, especially at 1B, shows much larger perplexity degradation.

\paragraph{XNLI task and language-family robustness.}
XNLI is used as a narrower robustness check for source retention and intervention robustness, not as a full replication of the NusaX multi-axis preservation suite.
Table~\ref{tab:xnli_core_results} summarizes the XNLI robustness check. Since XNLI starts from an English NLI competence-tuned checkpoint, the IndoNLU, SIB200, and Indonesian perturbation evaluations used for NusaX are not directly comparable. We use XNLI to test source retention and the RQ1 intervention finding on a harder task across two model families. Intervention results in Appendix~\ref{app:xnli_intervention_robustness} and Table~\ref{tab:xnli_faithfulness_knockout_combined} show that Circuit has the strongest knockout effects across both models. Faithfulness is lower than in NusaX, suggesting that NLI behavior is more distributed: the selected heads are load-bearing under intervention, but not sufficient alone to reconstruct the full behavior.

\begin{table}[ht]
\centering
\footnotesize
\setlength{\tabcolsep}{2.5pt}
\renewcommand{\arraystretch}{1.05}
\begin{tabular}{lcccc}
\toprule
\textbf{Method}
& \multicolumn{2}{c}{\textbf{Qwen-0.5B}}
& \multicolumn{2}{c}{\textbf{Llama-1B}} \\
\cmidrule(lr){2-3}
\cmidrule(lr){4-5}
& \textbf{Trans.} & \textbf{En}
& \textbf{Trans.} & \textbf{En} \\
\midrule
Comp. eval 
& 62.56
& \best{74.38}
& 44.39
& 47.00 \\

Circuit          
& 63.52
& 71.99
& \best{46.76}
& \best{47.04} \\

Avg. NC 
& \best{64.49}
& 73.29
& 46.73
& 45.82 \\

Full FT          
& 39.48
& 41.98
& 34.05
& 34.05 \\

LoRA             
& 63.09
& 71.63
& 46.34
& 46.43 \\
\bottomrule
\end{tabular}
\vspace{-0.6em}
\caption{
XNLI transfer and English retention. Scores are averaged over target languages, training sizes, and four seeds; Avg. NC averages Least-relevant, Near-zero, and Random NC. Full results are in Appendix~\ref{app:scalability_results}.
}
\label{tab:xnli_core_results}
\vspace{-1em}
\end{table}

Downstream, Full FT causes severe English forgetting, whereas sparse methods and LoRA remain closer to the competence-tuned checkpoint. On Qwen, Avg. non-circuit slightly exceeds Circuit on transfer, echoing the recruitment pattern from RQ2. Llama starts from weaker XNLI competence, so we treat it as a weak-competence stress test rather than a scale comparison. Overall, XNLI supports the preservation and intervention findings while showing that transfer can also benefit from non-circuit recruitment.

\section{Discussion and Conclusion}

This paper connects circuit discovery with controlled adaptation in natural-text classification. By adapting CD-T with label-balanced activation means and task-directional relevance scoring, we identify task-relevant attention-head circuits without manually constructed counterfactuals. These circuits show stronger faithfulness and knockout scores than matched sparse controls, making them useful intervention-supported targets for CT-SFT.

Our results suggest an editing--recruitment distinction. Circuit tuning updates heads identified by source-task intervention evidence, while non-circuit sparse updates can sometimes improve transfer by recruiting capacity outside the discovered circuit. Random NC is informative as a sparse-capacity control: it shows that restricted non-circuit subsets can support adaptation, but it is not a standalone selection rule because it is sampled from the complement of the discovered circuit. Its competitiveness is compatible with sparse-update findings and with broader evidence that transformer behavior can be distributed and partly compensatory. However, its weak knockout effects show that it is not the same kind of source-task intervention target as Circuit.

The baseline comparisons show why controlled adaptation matters. Full FT can improve transfer, but it causes severe behavioral drift. LoRA preserves sentiment-adjacent behavior better, but is weaker on broader Topic retention and label-preserving perturbations. The Llama-3B and XNLI extensions support the same framing: the selected circuit heads remain load-bearing under intervention, Full FT forgets strongly, and non-circuit gains recur as evidence of recruitment.

Overall, CT-SFT is not a method that consistently maximizes all metrics. Its value is that it provides a principled update location: a compact, knockout-supported set of task-relevant heads that can be adapted while limiting damage to existing behavior. When target accuracy alone matters, other sparse or PEFT methods may be competitive; when controlled low-resource adaptation matters, circuit targeting offers a more interpretable balance between transfer, preservation, and intervention evidence.

\section*{Limitations}

This work has several limitations regarding model competence, evaluation scope, baseline coverage, and interpretability granularity.

\paragraph{Dependence on checkpoint competence.}
CT-SFT assumes that the model already has usable source-task behavior before circuit discovery. As seen in the Llama-1B XNLI setting, if the competence-tuned checkpoint remains near chance, sparse adaptation cannot reliably construct the capability from scratch. CT-SFT should therefore be understood as an editing and routing technique that adapts an existing source-task capability, not as a method for creating that capability from scratch.

\paragraph{Scope of architectures and scales.}
Our experiments focus on relatively small decoder-only models: Llama-3.2-1B, Llama-3.2-3B, and Qwen2.5-0.5B. We do not evaluate encoder-only models or deployment-scale models. At larger scales, representations may become more distributed, which could change the redundancy and self-repair dynamics observed during circuit knockout.

\paragraph{Scope of tasks and evaluation axes.}
Our counterfactual-free CD-T adaptation relies on task-directional relevance scoring, which is most natural for tasks with discrete label tokens or compact output sets, such as classification, entity labeling, or controlled next-token prediction. Extending it to open-ended generation requires defining a suitable task direction over less constrained outputs. In addition, XNLI provides a task and language-family robustness check, but not a full replication of the NusaX multi-axis preservation suite. Thus, the strongest controlled-adaptation claims come from NusaX, while XNLI mainly supports source-retention and intervention findings.

\paragraph{Baseline and sparse-control scope.}
Our baseline comparisons isolate update location under a shared training protocol, rather than establishing dominance over all tuned baselines. Full FT and LoRA use the same data, seeds, epochs, learning rate, batch size, prompts, and evaluation pipeline as CT-SFT, but we do not perform extensive method-specific hyperparameter search. Random NC is also a strong sparse-capacity control in some settings, limiting any claim that circuit targeting is necessary for target accuracy alone. Our main claim is narrower: circuit targeting provides an intervention-supported update location with a more interpretable transfer--preservation tradeoff.

\paragraph{Intervention metrics.}
Faithfulness and knockout evaluate internal intervention behavior using label-token accuracy from a single forward pass, not autoregressively generated predictions. These metrics are useful for controlled circuit comparison but are not identical to downstream generation performance. Their absolute values may also depend on the intervention baseline, reference activations, and intervention sites.

\paragraph{Circuit granularity and interpretability.}
Our circuit granularity is restricted to attention heads. CT-SFT updates selected heads and LayerNorm, while MLPs, embeddings, the unembedding matrix, and unselected heads are frozen. This localization makes the update easier to analyze, but may miss task-relevant computation in MLPs, residual interactions, or finer-grained edges. The discovered circuits should therefore be interpreted as useful intervention-supported update targets, not exhaustive explanations of model behavior.

\section*{Use of AI Assistants}

We used AI assistants as writing and coding aids, including for improving clarity and grammar, restructuring text, drafting LaTeX/table snippets, debugging scripts, and refactoring small code fragments. All scientific content, experimental design, implementation decisions, analyses, interpretations, and reported results were produced and verified by the authors. AI assistants were not used to generate datasets, train models, or produce experimental results.

\bibliography{custom}
\clearpage
\appendix

\section{Experimental Details}
\label{app:experimental_details}

\subsection{Overview}
\label{app:experimental_overview}

This appendix provides the implementation, training, evaluation, circuit discovery, and update budget details for the experiments in the main paper. 
We separate the details into four parts for readability. 
Appendix~\ref{app:model_compute_setup} summarizes the model families and compute setup. 
Appendix~\ref{app:adaptation_eval_protocol} describes the adaptation and evaluation protocol shared across CT-SFT, sparse controls, Full FT, and LoRA. 
Appendix~\ref{app:training_discovery_protocol} describes circuit discovery, pruning, and the CT-SFT configuration. 
Appendix~\ref{app:update_budget} reports the effective updated-parameter budget for CT-SFT and the baselines.

Circuit discovery, circuit pruning, CT-SFT, and sparse-control evaluations were implemented in TransformerLens\footnote{\url{https://transformerlensorg.github.io/TransformerLens/}}. 
LoRA baselines were implemented using Hugging Face Transformers and PEFT. 
For each target language, all sparse-control methods are matched to the same post-pruning circuit head budget for each seed.

\subsection{Model and Compute Setup}
\label{app:model_compute_setup}

\begin{table}[ht]
\centering
\footnotesize
\setlength{\tabcolsep}{3pt}
\renewcommand{\arraystretch}{1.10}
\begin{tabular}{
  >{\raggedright\arraybackslash}p{0.38\linewidth}
  >{\raggedright\arraybackslash}p{0.52\linewidth}
}
\toprule
\textbf{Component} & \textbf{Setting} \\
\midrule
Models & Qwen2.5-0.5B, Llama-3.2-1B-Instruct, Llama-3.2-3B-Instruct \\
\addlinespace[1pt]
Approx. model sizes & 0.5B, 1B, and 3B parameters \\
\addlinespace[1pt]
Main model & Llama-3.2-1B-Instruct \\
\addlinespace[1pt]
Scalability model & Llama-3.2-3B-Instruct \\
\addlinespace[1pt]
XNLI robustness models & Qwen2.5-0.5B and Llama-3.2-1B \\
\addlinespace[1pt]
Compute infrastructure & SLURM-managed GPU cluster \\
\addlinespace[1pt]
GPU setup & Single GPU per run \\
\addlinespace[1pt]
GPU types & NVIDIA A40 \\
\addlinespace[1pt]
Memory consideration & Circuit discovery requires activation caching and is more memory-intensive than standard fine-tuning \\
\bottomrule
\end{tabular}
\caption{
Model sizes and computational setup.
}
\label{tab:model_compute_budget}
\end{table}

Table~\ref{tab:model_compute_budget} summarizes the model families and compute environment used in our experiments. 
The main experiments use Llama-3.2-1B-Instruct, while Llama-3.2-3B-Instruct is used for the NusaX scalability check. 
For XNLI, we evaluate both Llama-3.2-1B-Instruct and Qwen2.5-0.5B to test whether the source-retention and intervention patterns hold across model families. 
Each run uses a single GPU on a SLURM-managed cluster. 
Circuit discovery is more memory-intensive than standard fine-tuning because it requires activation caching for CD-T scoring and intervention evaluation.

\subsection{Adaptation and Evaluation Protocol}
\label{app:adaptation_eval_protocol}

Table~\ref{tab:training_eval_protocol} summarizes the shared adaptation and evaluation protocol. 
For NusaX, Indonesian is used as the source language for circuit discovery, and Acehnese, Buginese, Javanese, and Minangkabau are used as target languages. 
Each target language is adapted with $n \in \{25,50,75,100\}$ training examples across four seeds. 
All adaptation methods use the same target-language examples, prompt format, optimizer, learning rate, batch size, number of epochs, maximum sequence length, and evaluation pipeline. 
This shared protocol is intended to isolate the effect of update location under a common training setup, rather than to perform method-specific hyperparameter search.

\begin{table}[ht]
\centering
\footnotesize
\setlength{\tabcolsep}{3pt}
\renewcommand{\arraystretch}{1.10}
\begin{tabular}{
  >{\raggedright\arraybackslash}p{0.38\linewidth}
  >{\raggedright\arraybackslash}p{0.52\linewidth}
}
\toprule
\textbf{Component} & \textbf{Setting} \\
\midrule
Implementation framework & TransformerLens \\
\addlinespace[1pt]
Base model & Llama-3.2-1B-Instruct \\
\addlinespace[1pt]
Source language & Indonesian \\
\addlinespace[1pt]
Target languages & Acehnese, Buginese, Javanese, Minangkabau \\
\addlinespace[1pt]
Target training size & $n \in \{25,50,75,100\}$ \\
\addlinespace[1pt]
Seeds & $\{31,777,2025,12345\}$ \\
\addlinespace[1pt]
Optimizer & AdamW \\
\addlinespace[1pt]
Learning rate & $5 \times 10^{-5}$ \\
\addlinespace[1pt]
Training epochs & 5 \\
\addlinespace[1pt]
Batch size & 16 \\
\addlinespace[1pt]
Maximum sequence length & 128 \\
\addlinespace[1pt]
Training prompt style & Instruct format \\
\addlinespace[1pt]
Output generation & Maximum 3 new tokens \\
\addlinespace[1pt]
XNLI competence tuning size & 250 English examples \\
\bottomrule
\end{tabular}
\caption{Training and evaluation setup.}
\label{tab:training_eval_protocol}
\end{table}

\subsection{Circuit Discovery, Pruning, and CT-SFT Configuration}
\label{app:training_discovery_protocol}

\begin{figure}[ht]
    \centering
    \includegraphics[width=0.5\columnwidth]{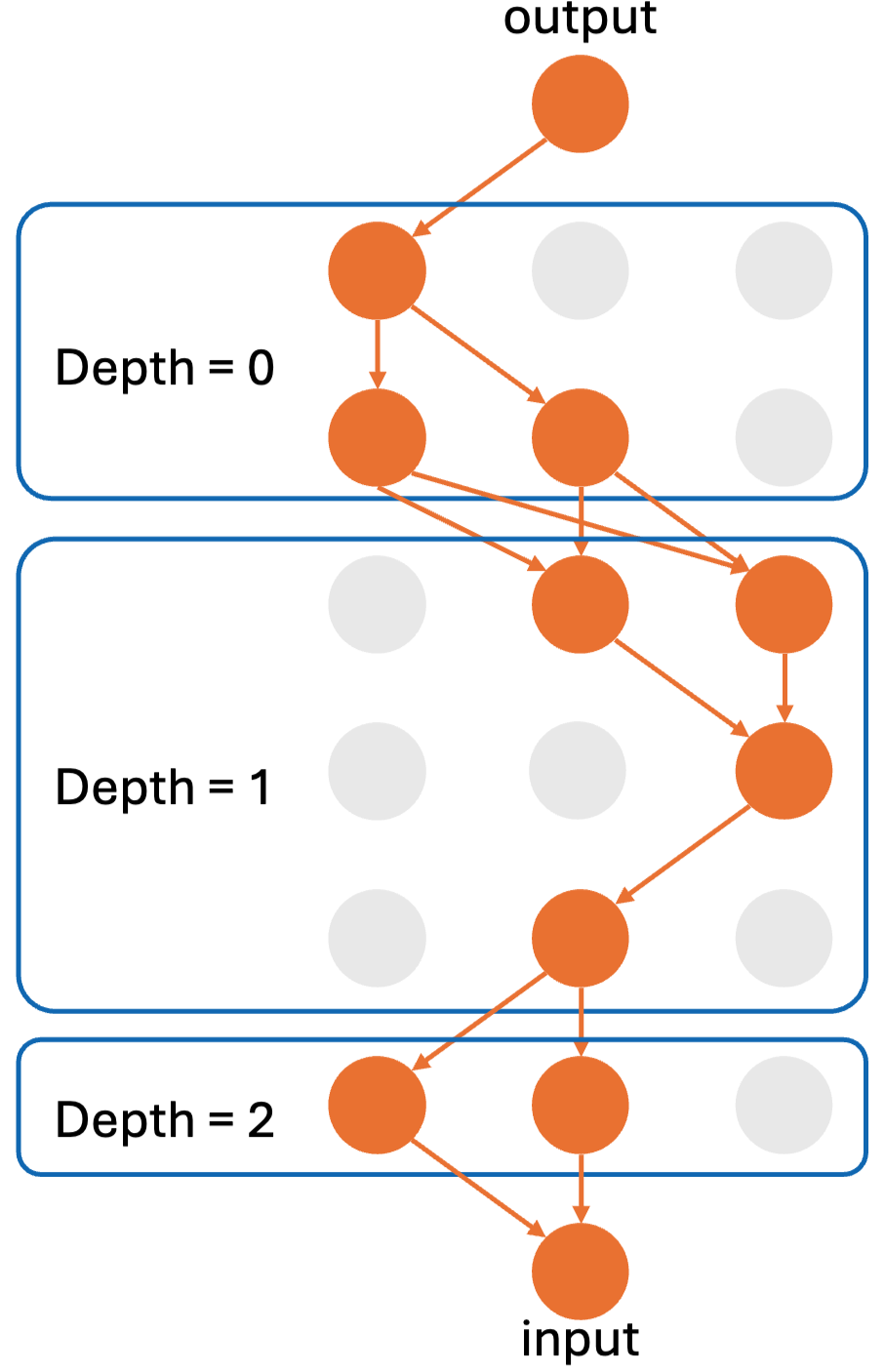}
    \caption{
    Iterative circuit construction. Beginning at the logit readout, we select the top-scoring heads, recursively expand backward by treating the selected heads as new targets, and form the final candidate circuit by taking the union of selected heads across depths.
    }
    \label{fig:iterative_circuit_construction}
    \vspace{-1em}
\end{figure}

Circuit discovery uses a label-balanced mean set and a separate correctly predicted discovery set from the source language. 
Heads are selected with task-directional relevance and expanded backward from the output decision. 
Figure~\ref{fig:iterative_circuit_construction} illustrates this iterative construction procedure: starting from the logit readout, the top-scoring heads are selected and then treated as new targets for the next backward expansion step. 
The final candidate circuit is the union of selected heads across depths.

After constructing the initial CD-T circuit, we prune the selected heads using a multi-pass faithfulness-guided greedy procedure. 
In each pass, candidate heads are considered in increasing order of absolute CD-T score. 
A head is removed if its removal improves the pruning objective while keeping the degradation of faithfulness within a fixed tolerance. 
We repeat this procedure for up to five passes, stopping early when a full pass removes no additional heads. 
The pruning objective is:
\[
\mathrm{Score}(C) = 0.7 \cdot \mathrm{Faith.}(C) + 0.3 \cdot \mathrm{Knockout}(C).
\]
Here, faithfulness measures the accuracy retained when only the circuit heads are kept and all non-circuit heads are mean-ablated. 
Knockout measures the drop in accuracy when the circuit heads are mean-ablated while all other heads remain active. 
We weight faithfulness higher than knockout to prioritize the standalone task competence of the discovered circuit, while using knockout as a secondary constraint to prune redundant heads. 
These weights were selected empirically based on preliminary experiments on the Indonesian validation set, where weighting knockout equally or higher led to overly aggressive pruning and degraded the circuit's standalone competence. 
Each accepted pruning step is constrained to have at most a 2-point faithfulness drop. 
The resulting pruned circuit is used for CT-SFT, and all sparse controls are matched to the final post-pruning head budget for each seed.

Table~\ref{tab:circuit_discovery_protocol} summarizes the circuit discovery, pruning, and CT-SFT configuration.

\begin{table}[ht]
\centering
\footnotesize
\setlength{\tabcolsep}{3pt}
\renewcommand{\arraystretch}{1.10}
\begin{tabular}{
  >{\raggedright\arraybackslash}p{0.38\linewidth}
  >{\raggedright\arraybackslash}p{0.52\linewidth}
}
\toprule
\textbf{Component} & \textbf{Setting} \\
\midrule
Mean set & 75 label-balanced Indonesian examples \\
\addlinespace[1pt]
Discovery set & 75 correctly predicted Indonesian examples \\
\addlinespace[1pt]
Scoring & Task-directional relevance \\
\addlinespace[1pt]
Circuit construction & Iterative backward expansion from output decision \\
\addlinespace[1pt]
Maximum discovery depth & Depths 0--2, corresponding to three backward expansion rounds \\
\addlinespace[1pt]
Loaded circuit depth for CT-SFT & Pruned cumulative circuit from depths 0--2 \\
\addlinespace[1pt]
Selection ratio & $p = 0.10$ per depth before pruning \\
\addlinespace[1pt]
Pruning objective & $0.7 \times$ faithfulness $+ 0.3 \times$ knockout \\
\addlinespace[1pt]
Faithfulness metric & Accuracy retained when non-circuit heads are mean-ablated \\
\addlinespace[1pt]
Knockout metric & Accuracy drop when circuit heads are mean-ablated \\
\addlinespace[1pt]
Pruning order & Lowest absolute CD-T score first \\
\addlinespace[1pt]
Pruning acceptance & Accept removal if objective improves and faithfulness drop is at most 2 points \\
\addlinespace[1pt]
Final circuit & Pruned circuit used for CT-SFT \\
\addlinespace[1pt]
CT-SFT trainable parameters & Selected circuit heads and LayerNorm \\
\addlinespace[1pt]
Frozen parameters & MLPs, embeddings, unembedding, and unselected heads \\
\addlinespace[1pt]
Sparse control budget & Matched to final post-pruning circuit size per seed \\
\bottomrule
\end{tabular}
\caption{Circuit discovery and CT-SFT setup.}
\label{tab:circuit_discovery_protocol}
\end{table}

\subsection{Effective Updated Parameter Budget}
\label{app:update_budget}

Table~\ref{tab:effective_update_budget} reports the effective updated-parameter budget for CT-SFT and the main baselines. For CT-SFT and matched sparse controls, these counts are not obtained by simply checking \texttt{requires\_grad}, because the implementation applies head-level gradient masking inside the training function. Accordingly, we report the effective update budget implied by the mask: selected attention-head Q/K/V/O slices plus LayerNorm. Full FT updates all model parameters, while LoRA updates only the q/v adapter parameters.

\begin{table*}[ht]
\centering
\footnotesize
\setlength{\tabcolsep}{5pt}
\renewcommand{\arraystretch}{1.08}
\begin{tabular}{@{}llcrr@{}}
\toprule
\textbf{Model} 
& \textbf{Method} 
& \textbf{Updated components} 
& \textbf{Eff. updated params} 
& \textbf{\% Full FT} \\
\midrule
Llama-1B 
& CT-SFT 
& 62.75 heads avg + LN 
& 29.10M 
& 2.355\% \\

Llama-1B 
& Matched sparse controls 
& 62.75 heads avg + LN 
& 29.10M 
& 2.355\% \\

Llama-1B 
& LoRA q/v $r=8$ 
& q/v all layers 
& 0.852M 
& 0.069\% \\

Llama-1B 
& Full FT 
& all parameters 
& 1,235.81M 
& 100\% \\

\midrule

Llama-3B 
& CT-SFT 
& 52.50 heads avg + LN 
& 77.44M 
& 2.410\% \\

Llama-3B 
& Matched sparse controls 
& 52.50 heads avg + LN 
& 77.44M 
& 2.410\% \\

Llama-3B 
& LoRA q/v $r=8$ 
& q/v all layers 
& 2.294M 
& 0.071\% \\

Llama-3B 
& Full FT 
& all parameters 
& 3,212.75M 
& 100\% \\
\bottomrule
\end{tabular}
\vspace{-0.6em}
\caption{
Effective updated-parameter counts for CT-SFT and baselines. 
CT-SFT and matched sparse controls update selected attention-head Q/K/V/O slices plus LayerNorm under a head-level gradient mask. 
LoRA updates q/v adapters only, and Full FT updates all parameters.
}
\label{tab:effective_update_budget}
\end{table*}

\subsection{Faithfulness and Knockout Metrics}
\label{app:faithfulness_knockout_metrics}

We evaluate discovered circuits using label-token intervention accuracy rather than generated accuracy. 
For each validation example, we append the gold label $y$ to the prompt $x$ and run a single forward pass. 
We then read the logits at the position immediately before the gold label token. 
The model prediction is the highest-logit label among the task-specific label set:
\[
\hat{y}
=
\arg\max_{y' \in \mathcal{Y}} z_{y'}(x),
\]
where $\mathcal{Y}$ is the task-specific label set and $z_{y'}(x)$ denotes the next-token logit assigned to label $y'$. 
For NusaX, $\mathcal{Y}=\{\texttt{negative},\texttt{neutral},\texttt{positive}\}$; for XNLI, $\mathcal{Y}=\{\texttt{entailment},\texttt{neutral},\texttt{contradiction}\}$.

Let $C$ be a selected set of attention heads. 
We use mean ablation at the attention-head output $z$: ablated heads are replaced by their mean activations computed over a label-balanced source-language mean set. 
Let $f_{\mathrm{full}}$ denote the unablated model, $f_{\mathrm{keep}\ C}$ denote the intervention where only heads in $C$ remain active and all non-selected heads are mean-ablated, and $f_{\mathrm{ablate}\ C}$ denote the intervention where heads in $C$ are mean-ablated while all other heads remain active.

Faithfulness is defined as the percentage of full-model label-token accuracy retained when only the selected heads are preserved:
\[
\mathrm{Faithfulness}(C)
=
100 \times
\frac{
\mathrm{Acc}(f_{\mathrm{keep}\ C})
}{
\mathrm{Acc}(f_{\mathrm{full}})
}.
\]

Knockout is defined as the relative drop in full-model label-token accuracy when the selected heads are ablated:
\[
\mathrm{Knockout}(C)
=
100 \times
\frac{
\mathrm{Acc}(f_{\mathrm{full}})
-
\mathrm{Acc}(f_{\mathrm{ablate}\ C})
}{
\mathrm{Acc}(f_{\mathrm{full}})
}.
\]

Both metrics are computed on the validation split. 
The mean set is sampled from the source-language training split using a label-balanced sampling procedure: Indonesian for NusaX and English for XNLI.

\subsection{Zero-Ablation Sensitivity}
\label{app:zero_ablation}

As a sensitivity check, we repeat the faithfulness and knockout evaluations using zero ablation rather than mean ablation. 
In this setting, ablated attention-head outputs are replaced with zero vectors rather than source-language mean activations. 
This intervention is harsher than mean ablation and is less aligned with CD-T, which decomposes activations around a mean/reference activation. 
Therefore, we use zero ablation only as a robustness check, not as the primary circuit-discovery or pruning criterion.

\begin{table}[ht]
\centering
\small
\setlength{\tabcolsep}{4pt}
\renewcommand{\arraystretch}{1.06}
\begin{tabular}{@{}llrr@{}}
\toprule
\textbf{Model} 
& \textbf{Method} 
& \textbf{Avg. Faith.} 
& \textbf{Avg. Knockout} \\
\midrule
Llama-1B & Circuit    & \best{56.73} & \best{34.55} \\
Llama-1B & Least-rel. & 47.20 & 8.32 \\
Llama-1B & Near-zero  & 43.94 & 3.81 \\
Llama-1B & Random NC  & 54.62 & 19.30 \\
\midrule
Llama-3B & Circuit    & 63.95 & \best{16.04} \\
Llama-3B & Least-rel. & 52.00 & 9.04 \\
Llama-3B & Near-zero  & \best{69.90} & 5.59 \\
Llama-3B & Random NC  & 52.00 & -1.80 \\
\bottomrule
\end{tabular}
\vspace{-0.6em}
\caption{
Aggregate zero-ablation sensitivity results. 
Faithfulness is retained full-model label-token accuracy when only the selected heads remain active and all other heads are zero-ablated. 
Knockout is the relative accuracy drop when the selected heads are zero-ablated. 
Scores are averaged across source and target languages.
}
\label{tab:zero_ablation_summary}
\end{table}

Table~\ref{tab:zero_ablation_summary} reports the aggregate zero-ablation results. 
Circuit remains strongest by average knockout in both Llama-1B and Llama-3B. 
This supports the conclusion that the selected heads are load-bearing under a second ablation baseline. 
Faithfulness is more variable under zero ablation: for example, Near-zero has high Llama-3B keep-only faithfulness but much lower knockout than Circuit. 

\begin{table*}[ht]
\centering
\setlength{\tabcolsep}{3pt}
\renewcommand{\arraystretch}{1.06}
\begin{tabular}{@{}llcccccc@{}}
\toprule
\textbf{Model} 
& \textbf{Method} 
& \textbf{Ind F/K} 
& \textbf{Ace F/K} 
& \textbf{Bug F/K} 
& \textbf{Jav F/K} 
& \textbf{Min F/K} 
& \textbf{Avg. F/K} \\
\midrule
Llama-1B & Circuit 
& \best{52.48}/\best{49.62} 
& \best{58.20}/\best{34.92} 
& 59.70/\best{15.32} 
& 53.35/\best{37.92} 
& \best{59.90}/\best{34.93} 
& \best{56.73}/\best{34.55} \\

Llama-1B & Least-rel. 
& 40.50/12.32 
& 46.98/7.33 
& 50.52/0.52 
& 44.58/16.23 
& 53.42/5.18 
& 47.20/8.32 \\

Llama-1B & Near-zero 
& 42.27/0.70 
& 43.98/7.75 
& 49.48/0.50 
& 40.83/5.83 
& 43.12/4.30 
& 43.94/3.81 \\

Llama-1B & Random NC 
& 43.33/29.55 
& 55.17/18.55 
& \best{67.85}/6.15 
& \best{54.58}/25.00 
& 52.15/17.23 
& 54.62/19.30 \\

\midrule
Llama-3B & Circuit 
& \best{57.42}/\best{30.85} 
& 70.67/\best{11.43} 
& 59.45/\best{3.33} 
& 65.50/\best{16.00} 
& 66.67/\best{18.62} 
& 63.95/\best{16.04} \\

Llama-3B & Least-rel. 
& 38.27/18.38 
& 54.90/3.23 
& 58.35/2.75 
& 55.50/11.50 
& 52.95/9.32 
& 52.00/9.04 \\

Llama-3B & Near-zero 
& 46.50/7.80 
& \best{75.00}/1.10 
& \best{93.35}/2.75 
& \best{67.50}/6.50 
& \best{67.12}/9.80 
& \best{69.90}/5.59 \\

Llama-3B & Random NC 
& 39.08/-5.85 
& 57.08/-1.65 
& 53.90/0.00 
& 57.00/-3.50 
& 52.95/1.98 
& 52.00/-1.80 \\
\bottomrule
\end{tabular}
\vspace{-0.6em}
\caption{
Per-language zero-ablation sensitivity results. 
Each cell reports faithfulness/knockout. 
Faithfulness is the retained full-model label-token accuracy under the keep-only-zero ablation, and knockout is the relative accuracy drop after zero-ablating the selected heads. 
Bold marks the best value within each model block and column.
}
\label{tab:zero_ablation_by_language}
\end{table*}

Table~\ref{tab:zero_ablation_by_language} gives the full per-language breakdown. 
The detailed results show the same qualitative pattern as the aggregate table. 
For Llama-1B, Circuit has the strongest average faithfulness and knockout, while Random NC is competitive in keep-only faithfulness but has substantially weaker knockout. 
For Llama-3B, Near-zero obtains the highest keep-only faithfulness on several target languages, but its knockout remains much smaller than Circuit's. 
This indicates that zero-ablation faithfulness can reflect redundancy or removal of interfering non-selected heads. In contrast, knockout more directly measures whether the selected heads are necessary for the task behavior.

\subsection{Evaluation Prompts}
\label{app:evaluation_prompts}

All downstream accuracy evaluations are generation-based. Intervention metrics use label-token accuracy, as defined in Appendix~\ref{app:faithfulness_knockout_metrics}. The model generates a short label string using greedy decoding, and the generated output is parsed with the same parser for CT-SFT, sparse controls, Full FT, and LoRA. NusaX transfer and source-language retention use at most three generated tokens, while fixed local evaluations such as IndoNLU EmoT and SIB200 use at most five generated tokens.

\begin{table*}[ht]
\centering
\footnotesize
\setlength{\tabcolsep}{4pt}
\renewcommand{\arraystretch}{1.12}
\begin{tabular*}{\textwidth}{
@{\extracolsep{\fill}}
p{0.20\textwidth}
p{0.18\textwidth}
p{0.20\textwidth}
p{0.14\textwidth}
p{0.22\textwidth}
@{}
}
\toprule
\textbf{Evaluation} & \textbf{Dataset} & \textbf{Language(s)} & \textbf{Size} & \textbf{Label space / criterion} \\
\midrule

Target transfer
& NusaX-Senti
& Acehnese, Buginese, Javanese, Minangkabau
& 400 per language
& negative, neutral, positive \\
\addlinespace[1pt]
Source-language retention
& NusaX-Senti
& Indonesian
& 400
& negative, neutral, positive \\
\addlinespace[1pt]
Behavior preservation
& IndoNLU EmoT
& Indonesian
& 400
& anger, fear, happy, love, sadness \\
\addlinespace[1pt]
Behavior preservation
& SIB200
& Indonesian
& 400
& topic labels \\
\addlinespace[1pt]
Controlled perturbation consistency
& NusaX-Senti-derived preserving pairs
& Indonesian
& 200 pairs
& prediction should remain unchanged; 70 negative, 70 positive, 60 neutral \\
\addlinespace[1pt]
Controlled perturbation consistency
& NusaX-Senti-derived flipping pairs
& Indonesian
& 200 pairs
& prediction should flip in the intended direction; 100 negative$\rightarrow$positive, 100 positive$\rightarrow$negative \\

\bottomrule
\end{tabular*}
\caption{
Evaluation datasets and controlled perturbation sets. All dataset-level evaluations are filtered to a maximum tokenized length of 128 tokens after prompt formatting, including the prompt and label text. We use label-balanced 400-example subsets for dataset-level evaluations, matching the NusaX-Senti test-set size. Controlled perturbation sets are derived from Indonesian NusaX-Senti test examples and evaluated at the pair level.
}
\label{tab:dataset_overview}
\end{table*}

\paragraph{NusaX sentiment classification.}
For controlled perturbation evaluation and for NusaX transfer/source evaluation when using the instruct prompt style, we use:

\begin{quote}
\small
Classify the sentiment of the following text.\\
Reply with exactly one word only: positive, negative, or neutral.\\
Do not explain.\\[0.5em]
Text: \{text\}\\
Sentiment:
\end{quote}

The valid labels are \texttt{positive}, \texttt{negative}, and \texttt{neutral}. In all NusaX evaluations, the generated text is parsed into one of these labels.

\paragraph{IndoNLU EmoT emotion detection.}
For IndoNLU EmoT, the evaluation uses the prompt stored in the fixed evaluation file. The model is evaluated with the same generation-based parser and the label set:

\begin{quote}
\small
Classify the emotion of the following Indonesian text.\\
Reply with exactly one label only: anger, fear, happy, love, sadness.\\
Do not explain.\\[0.5em]
Text: \{text\}\\
Emotion:
\end{quote}

\paragraph{SIB200 topic classification.}
For SIB200 topic classification, the evaluation uses the prompt stored in the fixed evaluation file. The valid labels are:

\begin{quote}
\small
Classify the topic of the following text.\\
Reply with exactly one label only: science/technology, travel, politics, sports, health, entertainment, geography.\\
Do not explain.\\[0.5em]
Text: \{text\}\\
Topic:
\end{quote}

\paragraph{Controlled perturbation evaluation.}
For label-preserving and label-flipping perturbation pairs, we use the NusaX sentiment prompt above for both the original and perturbed sentences. Label-preserving consistency is counted as successful when the prediction remains unchanged across the pair. Label-flipping consistency is counted as successful when the original prediction matches the original label and the perturbed prediction matches the intended flipped label.

\subsection{Evaluation Datasets and Examples}
\label{app:dataset_details}

This section summarizes the evaluation datasets used in our experiments and provides examples from the actual evaluation sets. Table~\ref{tab:dataset_overview} gives an overview of all evaluation settings, including the dataset, language, size, and label space or success criterion. We group the evaluations into target transfer, behavior preservation, and controlled perturbation consistency. Target transfer and source-language retention use NusaX-Senti, while IndoNLU EmoT and SIB200 are used to evaluate whether adaptation preserves related and less-related classification behavior. The perturbation sets are derived from Indonesian sentiment examples and are evaluated at the pair level.

\paragraph{Evaluation set construction.}
All evaluation examples were obtained from the official Hugging Face dataset releases. Before evaluation, we applied the same prompt template used by the model and filtered out examples whose tokenized length exceeded 128 tokens, including the prompt and label text. For each evaluation dataset, we then constructed label-balanced subsets of 400 examples, matching the size of the NusaX-Senti test set used in the main transfer experiments. This filtering and balancing procedure was applied consistently to NusaX-Senti, IndoNLU EmoT, and SIB200.

For controlled perturbation evaluation, we constructed Indonesian sentiment pairs from the NusaX-Senti test set. The label-flipping set contains 200 pairs: 100 negative-to-positive and 100 positive-to-negative. The label-preserving set also contains 200 pairs: 70 negative, 70 positive, and 60 neutral pairs. Label-preserving pairs are used to test prediction invariance, while label-flipping pairs are used to test whether predictions change in the intended sentiment direction.

\paragraph{NusaX-Senti.}
Table~\ref{tab:nusax_examples} shows a parallel NusaX-Senti example across Indonesian and the four target languages used in our transfer experiments.

\begin{table}[ht]
\centering
\footnotesize
\setlength{\tabcolsep}{3pt}
\renewcommand{\arraystretch}{1.12}
\begin{tabular}{
  >{\raggedright\arraybackslash}p{0.24\linewidth}
  >{\raggedright\arraybackslash}p{0.54\linewidth}
  >{\raggedright\arraybackslash}p{0.14\linewidth}
}
\toprule
\textbf{Lang.} & \textbf{Example} & \textbf{Label} \\
\midrule
Indonesian 
& \textit{Kecewa saya dengan provider ini, sampai kesal, marah, emosi, saya keluar, gara gara, sinyal yang tidak benar.} 
& negative \\

Acehnese 
& \textit{Keucewa lon keu provider nyoe, luwat, beungeh, emosi, lon teubiet, gara-gara sinyal nyang hana pah.} 
& negative \\

Buginese 
& \textit{Masara nyawa ka mitai iyewe provider e, lettu' ka mapella, macai, mappiddara, iya massu, nasaba', sinyal e de na tuju.} 
& negative \\

Javanese 
& \textit{Gela aku karo provider iki, nganti sebel, nesu, emosi, aku metu, gara gara, sinyal sing ora bener.} 
& negative \\

Minangkabau 
& \textit{Kecewa awak samo provider ko, sampai kesal, berang, emosi, awak kalua, dek karano, sinyal nan indak batua.} 
& negative \\

\bottomrule
\end{tabular}
\caption{
Actual NusaX-Senti examples from the Indonesian source language and four target languages used for transfer evaluation.
}
\label{tab:nusax_examples}
\end{table}

\paragraph{IndoNLU EmoT.}
Table~\ref{tab:emot_examples} shows examples from the IndoNLU EmoT evaluation set. 
This task is used to evaluate related-task retention because emotion detection and sentiment classification both involve affective language understanding. 
Unlike NusaX-Senti, however, IndoNLU EmoT requires the model to distinguish finer-grained emotional categories such as anger, fear, happiness, love, and sadness rather than only positive, negative, and neutral sentiment. 
We therefore use it as a related but non-identical preservation test: a model that adapts to target-language sentiment while preserving IndoNLU EmoT performance is less likely to have damaged its broader affective-language representations.

\begin{table}[ht]
\centering
\footnotesize
\setlength{\tabcolsep}{3pt}
\renewcommand{\arraystretch}{1.12}
\begin{tabular}{
  >{\raggedright\arraybackslash}p{0.70\linewidth}
  >{\raggedright\arraybackslash}p{0.20\linewidth}
}
\toprule
\textbf{Example} & \textbf{Label} \\
\midrule
\textit{bikin orang kesal saja}
& anger \\

\textit{Ya.....Allah..smoga 2019 tdk ada lagi 2 manusia super hero di DPR}
& fear \\

\textit{gp sudah hitam manis laku pula dari pada cuma ngaku2in pacar orang yeh}
& happy \\

\textit{Aku suka caramu menyukaiku :')}
& love \\

\textit{Buka puasa cuman minum sama makan gorengan, trus tengah malam makan bakso cabe rawitnya 3 sendok makan lebih, abis itu perut langsung melilit}
& sadness \\

\bottomrule
\end{tabular}
\caption{
Actual examples from the IndoNLU EmoT evaluation set.
}
\label{tab:emot_examples}
\end{table}

\paragraph{SIB200 topic classification.}
Table~\ref{tab:sib200_examples} shows examples from the Indonesian SIB200 topic classification evaluation set. This task is used to evaluate the preservation of less-related behavior beyond sentiment and affective classification. Unlike NusaX-Senti and IndoNLU EmoT, SIB200 requires the model to assign texts to broad semantic topics such as science/technology, sports, health, politics, and travel. We use it as a stricter preservation test: strong SIB200 retention suggests that target-language sentiment adaptation has not substantially damaged more general Indonesian classification behavior. This helps distinguish methods that preserve only sentiment-adjacent behavior from methods that maintain broader task competence after adaptation.

\begin{table}[ht]
\centering
\footnotesize
\setlength{\tabcolsep}{3pt}
\renewcommand{\arraystretch}{1.12}
\begin{tabular}{
  >{\raggedright\arraybackslash}p{0.70\linewidth}
  >{\raggedright\arraybackslash}p{0.30\linewidth}
}
\toprule
\textbf{Example} & \textbf{Label} \\
\midrule
\textit{Ia melakukan pembuatan bel pintu dengan teknologi WiFi, katanya.}
& science/\newline technology \\

\textit{Baik golf maupun rugbi akan kembali ke Olimpiade.}
& sports \\

\textit{Ia kemudian dipindahkan ke Rumah Sakit Addenbrooke di Cambridge.}
& health \\

\textit{Ini berlawanan dengan laporan sebelumnya, yang menyatakan bahwa membatalkan pemilu ulang berarti menentang konstitusi.}
& politics \\

\textit{Turkish Airlines terbang ke 39 tujuan di 30 negara Afrika pada 2014.}
& travel \\

\bottomrule
\end{tabular}
\caption{
Actual examples from the Indonesian SIB200 topic classification evaluation set.
}
\label{tab:sib200_examples}
\end{table}

\begin{table*}[ht]
\centering
\footnotesize
\setlength{\tabcolsep}{4pt}
\renewcommand{\arraystretch}{1.12}
\begin{tabular*}{\textwidth}{
@{\extracolsep{\fill}}
p{0.09\textwidth}
p{0.35\textwidth}
p{0.35\textwidth}
p{0.18\textwidth}
@{}
}
\toprule
\textbf{Type} & \textbf{Original} & \textbf{Perturbed} & \textbf{Expected} \\
\midrule

Preserving
& \textit{Karena lapar, saya pesan indomie.}
& \textit{Karena lapar, aku pesan indomie.}
& neutral $\rightarrow$ neutral \\

Preserving
& \textit{Belanja di Giant diskon 20\% dengan kartu kredit bca.}
& \textit{Belanja di Giant diskon 20\% dengan kartu kredit bni.}
& neutral $\rightarrow$ neutral \\

Preserving
& \textit{Pagi ini lupa lepas helm abang gojek setelah turun dari motor gara-gara mengejar bus}
& \textit{Siang ini lupa lepas helm abang maxim setelah turun dari motor gara-gara mengejar travel}
& neutral $\rightarrow$ neutral \\

Preserving
& \textit{Salah satu mal dengan konsep yang cukup unik. Banyak pilihan restoran sehingga cocok untuk menghabiskan waktu bersama teman-teman.}
& \textit{Salah satu mal dengan konsep yang cukup unik. Banyak pilihan kafe sehingga cocok untuk menghabiskan waktu bersama sahabat-sahabat.}
& positive $\rightarrow$ positive \\

Preserving
& \textit{Parkiran susah bener sampai berputar 3x. Saran: Parkir di dekat hotel saja ,tlalujalan kaki. Restoran di area perumahan.}
& \textit{Parkiran susah bener sampai berputar 3x. Saran: Parkir di dekat villa saja terus jalan kaki. Kafe di area perumahan.}
& negative $\rightarrow$ negative \\

\midrule

Flipping
& \textit{Saya tidak kecewa dengan produk apple}
& \textit{Saya sangat kecewa dengan produk apple}
& positive $\rightarrow$ negative \\

Flipping
& \textit{Harga selangit porsi seiprit tidak janji lagi saya datang untuk yang kedua kali. Masakannya tidak beda jauh sama ema saya dong!}
& \textit{Harga bersahabat porsi melimpah, pasti saya datang untuk yang kedua kali. Masakannya jauh lebih enak dari ema saya dong!}
& negative $\rightarrow$ positive \\

Flipping
& \textit{Macet di mana-mana kalau lagi liburan}
& \textit{Lancar di mana-mana walaupun lagi liburan}
& negative $\rightarrow$ positive \\

Flipping
& \textit{Rumah makan nusa indah sekarang mengecewakan ah. Kotor, orang didiamkan saja merokok di ruangan ber ac, makanan lama datangnya.}
& \textit{Rumah makan nusa indah sekarang memuaskan ah. Bersih, orang dilarang tegas merokok di ruangan ber ac, makanan sangat cepat datangnya.}
& negative $\rightarrow$ positive \\

Flipping
& \textit{Ini adalah sop stengkel terenak di aceh tamiang.}
& \textit{Ini adalah sop stengkel terburuk di aceh tamiang.}
& positive $\rightarrow$ negative \\

\bottomrule
\end{tabular*}
\caption{
Actual examples from the Indonesian controlled perturbation sets. Label-preserving pairs test prediction invariance, while label-flipping pairs test whether predictions change in the intended sentiment direction.
}
\label{tab:perturbation_examples}
\end{table*}

\paragraph{Controlled perturbation examples.}
Table~\ref{tab:perturbation_examples} shows examples from the Indonesian controlled perturbation sets. Label-preserving pairs keep the original sentiment label unchanged and are counted as successful when the model prediction remains unchanged. Label-flipping pairs intentionally alter sentiment-bearing words and are counted as successful when the original prediction matches the original label and the perturbed prediction matches the intended flipped label.

\subsection{LoRA Baseline}
\label{app:lora_baseline}

We include LoRA as a standard reference for parameter-efficient fine-tuning. Unlike CT-SFT, LoRA does not use circuit information and is not intended as a mechanism-localized intervention. We report it as an additional adaptation baseline, rather than as part of the matched sparse-control comparison.

\paragraph{Training setup.}
For each target language and training size, LoRA is trained on the same target-language examples used by the other adaptation methods. Specifically, for transfer from Indonesian to a target language $L$, we train LoRA on $n \in \{25, 50, 75, 100\}$ examples from $L$, using the same random seeds, data subsampling strategy, prompt format, batch size, number of epochs, and learning rate as the other fine-tuning baselines.

We use the Hugging Face PEFT implementation with LoRA applied to the query and value projection modules, \texttt{q\_proj} and \texttt{v\_proj}. The LoRA rank is $r=8$, the scaling factor is $\alpha=16$, and dropout is set to $0.05$. We train for 5 epochs using AdamW with learning rate $5 \times 10^{-5}$, batch size 16, and maximum sequence length 128. During training, the base model parameters are frozen and only the LoRA adapter parameters are updated.

\paragraph{Evaluation protocol.}
After training, we merge the LoRA adapter into the base model using the PEFT \texttt{merge\_and\_unload()} procedure. The merged model is then evaluated using the same generation-based evaluation pipeline as CT-SFT, Full FT, and the sparse controls. This ensures that LoRA uses the same prompts, decoding procedure, parsing functions, and held-out evaluation examples as the other methods.

We evaluate LoRA on target-language NusaX sentiment accuracy, Indonesian source-language retention, IndoNLU EmoT retention, SIB200 topic retention, and Indonesian controlled perturbation consistency using label-preserving and label-flipping pairs. The perturbation metrics are reported as pair-level success rates. Temporary adapter and merged-model artifacts are deleted after evaluation; only the evaluation outputs and summary files are retained.

\paragraph{Role of the baseline.}
LoRA provides a practical PEFT reference, but it does not test the same mechanistic hypothesis as CT-SFT. CT-SFT asks whether updating an intervention-supported source-task circuit gives a controlled adaptation target, while LoRA applies low-rank updates to predefined projection modules independent of circuit discovery. Accordingly, LoRA is used to contextualize CT-SFT, while the main analysis focuses on circuit updates and matched non-circuit sparse controls.

\subsection{Artifacts and Licenses}
\label{app:artifacts_licenses}

We use publicly available research artifacts: pretrained language models, public benchmark datasets, and open-source software libraries. We use these artifacts for research evaluation only and in accordance with their intended research use. Dataset and model artifacts are cited in the main paper, and implementation libraries are listed in Appendix~\ref{app:experimental_details}.

\section{Additional Llama-1B NusaX Results}
\label{app:additional_results}

This appendix provides additional analyses for the main Llama-3.2-1B NusaX setting. 
The main paper reports the core transfer, preservation, and intervention results, while this appendix gives more detailed views of the same experimental setting. 

We organize the analyses around five complementary questions. 
First, we examine whether CT-SFT's gains are driven primarily by the selected attention heads or by LayerNorm. 
Second, we summarize how Circuit and matched sparse controls trade off the six downstream evaluation axes. 
Third, we show how target-language transfer varies with the number of target-language training examples. 
Fourth, we compare Circuit with LoRA in more detail to separate mechanism-guided adaptation from a same-data PEFT reference. 
Finally, we examine fixed-task retention trends on SIB200 Topic and IndoNLU EmoT to show how target-language adaptation affects broader Indonesian behavior across training sizes.

\begin{table*}[t]
\centering
\footnotesize
\setlength{\tabcolsep}{5pt}
\renewcommand{\arraystretch}{1.08}
\begin{tabular}{@{}lcccccc@{}}
\toprule
\textbf{Method} 
& \textbf{Ind/Source} 
& \textbf{Ace} 
& \textbf{Bug} 
& \textbf{Jav} 
& \textbf{Min} 
& \textbf{Target Avg.} \\
\midrule
Pre-adapt. 
& 73.50 & 52.25 & 39.00 & 56.00 & 52.00 & 49.81 \\
LN-only 
& \best{73.84} & \best{53.83} & 39.45 & 57.28 & 52.23 & 50.70 \\
Heads-only no-LN 
& 71.94 & 50.22 & 42.75 & \best{63.86} & 60.05 & 54.22 \\
CT-SFT heads+LN 
& 69.55 & 50.02 & \best{42.86} & 63.48 & \best{61.27} & \best{54.41} \\
\bottomrule
\end{tabular}
\vspace{-0.6em}
\caption{
Per-language Llama-1B LayerNorm ablation results.
Ind/Source is Indonesian source-language retention; Target Avg. is the average of Acehnese, Buginese, Javanese, and Minangkabau target transfer.
LN-only updates only LayerNorm/RMSNorm parameters, Heads-only no-LN updates only selected attention heads, and CT-SFT heads+LN is the original CT-SFT setting.
}
\label{tab:layernorm_ablation_by_language}
\end{table*}

\subsection{LayerNorm Ablation Details}
\label{app:layernorm_details}

The main text reports the aggregate LayerNorm ablation in Table~\ref{tab:layernorm_ablation}. 
Here, we provide two additional views. 
First, Table~\ref{tab:layernorm_ablation_by_language} breaks the result down by source and target language. 
Second, Table~\ref{tab:heads_only_sparse_controls} compares heads-only no-LN updates across Circuit and matched sparse controls. 
Together, these results support the interpretation that LayerNorm alone is not the primary driver of CT-SFT's transfer and Topic gains.

The per-language results show the same pattern as the aggregate table. 
LN-only best preserves Indonesian source performance but provides little target-language adaptation. 
In contrast, heads-only no-LN and CT-SFT heads+LN produce much larger target-language gains, especially for Buginese, Javanese, and Minangkabau. 
This indicates that the selected attention heads carry most of the adaptation capacity, while LayerNorm provides an auxiliary stabilizing or complementary update.

When LayerNorm is frozen for all sparse methods, Circuit remains strongest on the balanced controlled-adaptation profile: it gives the best transfer, source retention, Emotion retention, and label-preserving consistency. 
Near-zero remains strongest on label-flipping consistency, consistent with the plasticity interpretation discussed in the main text. 
Random NC remains competitive, but unlike Circuit, it is not selected by source-task intervention evidence.

\begin{table}[ht]
\centering
\small
\setlength{\tabcolsep}{3pt}
\renewcommand{\arraystretch}{1.06}
\begin{tabular}{@{}lrrrr@{}}
\toprule
\textbf{Eval.} 
& \textbf{Circuit} 
& \textbf{Least-rel.} 
& \textbf{Near-zero} 
& \textbf{Random NC} \\
\midrule
Transfer   
& \best{54.22} & 52.66 & 51.48 & 53.44 \\
Source     
& \best{71.94} & 71.88 & 70.13 & 71.82 \\
Emotion    
& \best{41.30} & 34.62 & 40.57 & 36.71 \\
Topic      
& 67.48 & \best{69.01} & 56.50 & 64.39 \\
Pres. Pair 
& \best{64.10} & 62.93 & 62.55 & 63.09 \\
Flip Pair  
& 85.81 & 82.45 & \best{86.13} & 85.45 \\
\bottomrule
\end{tabular}
\vspace{-0.6em}
\caption{
Aggregate Llama-1B heads-only no-LN sparse-control results.
All methods update only attention heads and freeze LayerNorm/RMSNorm.
Scores are averaged over four target-language training sizes and four seeds. Higher is better.
}
\label{tab:heads_only_sparse_controls}
\end{table}

\subsection{Sparse-Control Rank Summary}
\label{app:sparse_rank_summary}

Table~\ref{tab:sparse_rank_summary} summarizes the metric-wise ranks of the sparse methods in the main Llama-1B NusaX experiment. 
We include this table as a descriptive overview of the downstream trade-offs across transfer, source retention, related-task retention, less-related-topic retention, and perturbation consistency. 
Circuit has the best average rank, but we do not use this summary alone as evidence that circuit targeting is practically necessary. 
Rather, it complements the intervention results in Table~\ref{tab:faithfulness_knockout_by_model_lang} and the preservation analyses in the main text.

\begin{table}[ht]
\centering
\small
\setlength{\tabcolsep}{3.5pt}
\renewcommand{\arraystretch}{1.06}
\begin{tabular}{@{}lcccc@{}}
\toprule
\textbf{Metric} 
& \textbf{Circuit} 
& \textbf{Least-rel.} 
& \textbf{Near-zero} 
& \textbf{Random NC} \\
\midrule
Transfer   
& \best{54.40 (1)} & 53.20 (3) & 50.50 (4) & 53.60 (2) \\
Source     
& 69.50 (3) & 69.70 (2) & 65.50 (4) & \best{70.30 (1)} \\
Emotion    
& \best{40.20 (1)} & 32.90 (4) & 39.40 (2) & 36.10 (3) \\
Topic      
& \best{69.30 (1)} & 69.20 (2) & 58.70 (4) & 64.30 (3) \\
Pres. Pair 
& 64.70 (2) & 64.10 (3) & 62.00 (4) & \best{64.90 (1)} \\
Flip Pair  
& 79.50 (2) & 73.80 (4) & \best{81.90 (1)} & 79.30 (3) \\
\midrule
Avg. rank  
& \best{1.67} & 3.00 & 3.17 & 2.17 \\
\bottomrule
\end{tabular}
\vspace{-0.6em}
\caption{
Metric-wise sparse-control ranks on the main Llama-1B NusaX setting.
Numbers in parentheses indicate rank among sparse methods for each metric.
Average rank is reported only as a descriptive summary and is not used as the primary evidence for practical necessity.
}
\label{tab:sparse_rank_summary}
\end{table}

\begin{figure*}[ht]
    \centering
    \includegraphics[width=0.85\linewidth]{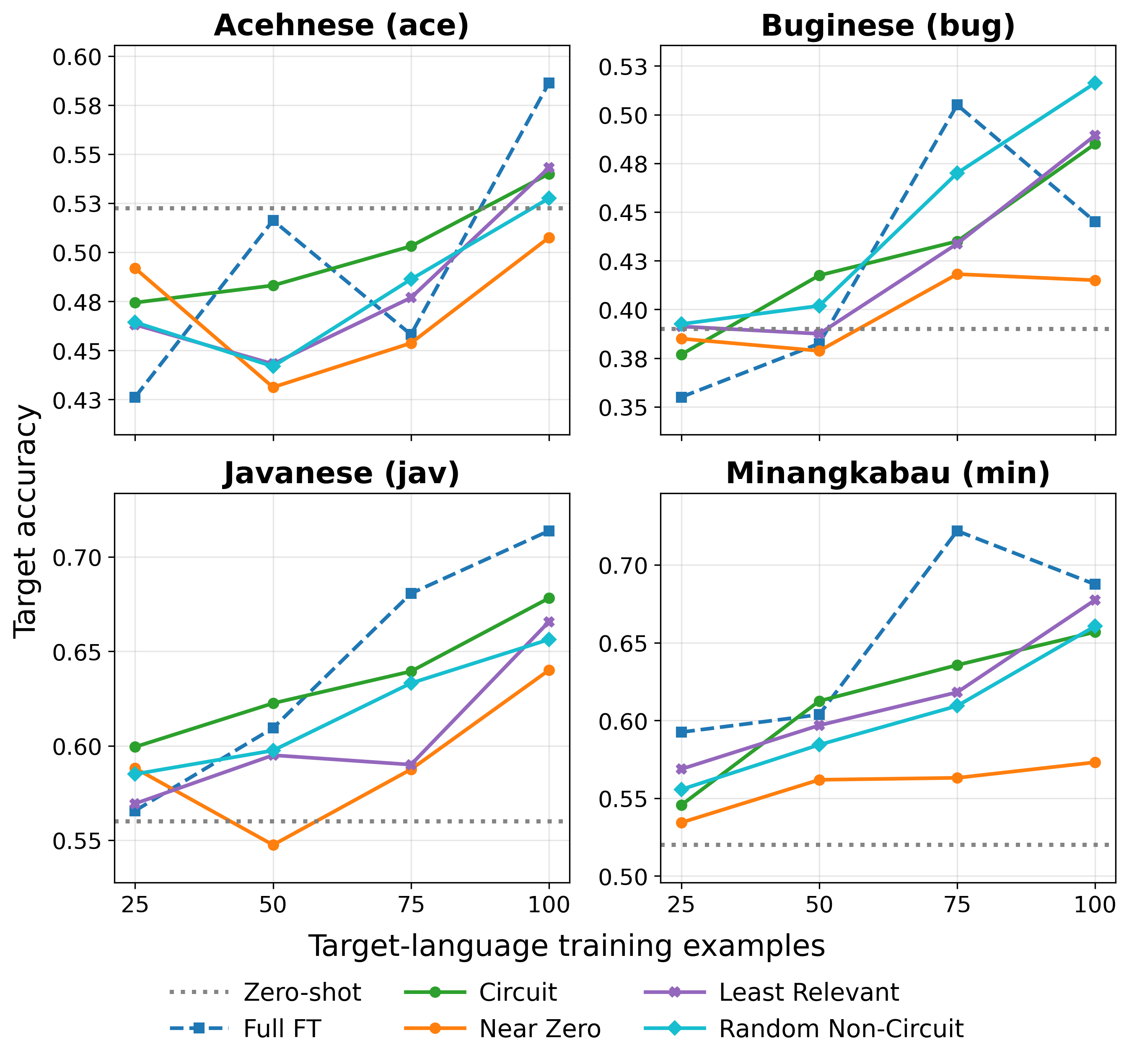}
    \caption{
    Target-language transfer across training sizes for Llama-3.2-1B-Instruct.
    Each panel corresponds to one target language, with zero-shot and Full FT included as reference baselines.
    }
    \label{fig:target_transfer_trends}
\end{figure*}

\subsection{Target Transfer Trends by Training Size}
\label{app:target_transfer_trends}

Figure~\ref{fig:target_transfer_trends} shows target-language transfer accuracy across the four low-resource training sizes, $n \in \{25,50,75,100\}$, for each target language. 
This complements the aggregate transfer scores in Table~\ref{tab:transfer_preservation_by_lang_llama1b} by showing whether each method's behavior is stable across data sizes rather than driven by a single training budget.

Overall, Circuit remains competitive across training sizes and target languages, while non-circuit sparse controls can also improve with additional target-language data. 
The trends support the interpretation in RQ2: target transfer can arise both from updating the discovered source-task circuit and from recruiting non-circuit capacity. 
However, because transfer accuracy alone does not distinguish controlled editing from uncontrolled recruitment, these curves should be interpreted alongside the retention, perturbation, and intervention results reported in the main text.

\subsection{Circuit and LoRA Details}
\label{app:lora_circuit_details}

Table~\ref{tab:lora_circuit_by_language} provides the per-target-language breakdown for Circuit and LoRA in the main Llama-1B NusaX setting. 
This complements the aggregate reference-baseline comparison in Table~\ref{tab:reference_baselines} by showing whether the aggregate trends are consistent across Acehnese, Buginese, Javanese, and Minangkabau. 
LoRA is included as a same-data and same-protocol PEFT reference: it uses the same target-language examples, training sizes, random seeds, epochs, learning rate, batch size, prompt format, decoding setup, parsing function, and held-out evaluation pipeline as the other adaptation methods. 
However, unlike CT-SFT, LoRA does not use circuit information or select update locations based on source-task intervention evidence.

\begin{table*}[ht]
\centering
\footnotesize
\setlength{\tabcolsep}{4pt}
\renewcommand{\arraystretch}{1.06}
\begin{tabular*}{\textwidth}{@{\extracolsep{\fill}}llccccc@{}}
\toprule
\textbf{Method} & \textbf{Evaluation}
& \textbf{Ace} & \textbf{Bug} & \textbf{Jav} & \textbf{Min} & \textbf{Avg.} \\
\midrule

Circuit 
& Transfer & 50.00 & \best{42.90} & \best{63.50} & \best{61.30} & \best{54.40} \\
& Source & 69.20 & 68.40 & 70.30 & 70.30 & 69.50 \\
& Emotion & 40.10 & 39.10 & \best{41.40} & 40.00 & 40.20 \\
& Topic & \best{68.40} & \best{69.40} & \best{69.60} & \best{69.60} & \best{69.30} \\
& Pres. Pair & \best{64.30} & \best{63.80} & \best{65.70} & \best{65.00} & \best{64.70} \\
& Flip Pair & 79.90 & 78.40 & 79.90 & 79.70 & 79.50 \\

\midrule

LoRA
& Transfer 
& \best{\scoreDelta{53.28}{+3.28}} 
& \scoreDelta{39.20}{-3.70} 
& \scoreDelta{56.58}{-6.92} 
& \scoreDelta{52.22}{-9.08} 
& \scoreDelta{50.32}{-4.08} \\

& Source 
& \best{\scoreDelta{72.95}{+3.75}} 
& \best{\scoreDelta{73.03}{+4.63}} 
& \best{\scoreDelta{72.97}{+2.67}} 
& \best{\scoreDelta{72.98}{+2.68}} 
& \best{\scoreDelta{72.98}{+3.48}} \\

& Emotion 
& \best{\scoreDelta{41.06}{+0.96}} 
& \best{\scoreDelta{40.95}{+1.85}} 
& \scoreDelta{41.20}{-0.20} 
& \best{\scoreDelta{41.08}{+1.08}} 
& \best{\scoreDelta{41.07}{+0.87}} \\

& Topic 
& \scoreDelta{48.27}{-20.13} 
& \scoreDelta{47.81}{-21.59} 
& \scoreDelta{49.25}{-20.35} 
& \scoreDelta{48.72}{-20.88} 
& \scoreDelta{48.51}{-20.79} \\

& Pres. Pair 
& \scoreDelta{63.06}{-1.24} 
& \scoreDelta{62.94}{-0.86} 
& \scoreDelta{63.03}{-2.67} 
& \scoreDelta{63.12}{-1.88} 
& \scoreDelta{63.04}{-1.66} \\

& Flip Pair 
& \best{\scoreDelta{82.38}{+2.48}} 
& \best{\scoreDelta{82.47}{+4.07}} 
& \best{\scoreDelta{82.62}{+2.72}} 
& \best{\scoreDelta{82.69}{+2.99}} 
& \best{\scoreDelta{82.54}{+3.04}} \\

\bottomrule
\end{tabular*}
\vspace{-0.8em}
\caption{
Circuit and LoRA results by target language, averaged across training sizes and seeds.
Parentheses indicate the signed percentage-point difference between LoRA and Circuit.
Bold marks the better score between Circuit and LoRA for each evaluation and language.
}
\label{tab:lora_circuit_by_language}
\end{table*}

The per-language results show a consistent tradeoff between sentiment-adjacent preservation and broader controlled adaptation. 
LoRA preserves Indonesian source performance, IndoNLU Emotion retention, and label-flipping consistency better than Circuit. 
This suggests that LoRA is a strong preservation-oriented baseline for behavior that closely aligns with the original sentiment objective. 
In particular, its stronger label-flipping consistency indicates that low-rank updates can remain responsive to explicit sentiment-changing cues. 
However, this advantage does not extend uniformly to all preservation axes.

Circuit is stronger on average target transfer, SIB200 Topic retention, and label-preserving perturbation consistency. 
The Topic gap is especially consistent across all target languages, indicating that LoRA's preservation is narrower than Circuit's broader controlled-adaptation profile. 
This is important because SIB200 is less directly tied to sentiment than IndoNLU Emotion, so Topic retention provides a stricter test of whether adaptation disrupts broader classification behavior. 
Similarly, stronger label-preserving consistency suggests that Circuit better maintains prediction invariance under meaning-preserving perturbations, rather than only responding flexibly to explicit sentiment flips. 

These results support the interpretation in the main text: LoRA is a strong preservation-oriented PEFT reference, especially for behavior close to the sentiment objective, but it does not provide the same broader preservation profile as Circuit. 
Circuit is not uniformly better on every metric, and LoRA is clearly competitive on source retention, Emotion retention, and label-flipping consistency. 
However, Circuit better preserves less-related Topic behavior and label-preserving consistency while remaining competitive for target transfer. 
Thus, the difference between Circuit and LoRA is not simply about target-language accuracy; it reflects different preservation profiles under the same data and training protocol.

This distinction also clarifies the role of LoRA in our evaluation. 
LoRA is a useful practical baseline, but it does not test the same mechanistic hypothesis as CT-SFT. 
CT-SFT asks whether a source-task circuit identified by intervention evidence can serve as a controlled update target. 
LoRA instead applies low-rank updates to predefined projection modules independent of circuit discovery. 
The per-language comparison therefore supports the main claim that mechanism-guided updates offer a different kind of control: not universal dominance on every score, but a more interpretable tradeoff between transfer, broad behavior preservation, and perturbation consistency.

\subsection{Fixed-Task Retention by Target Language and Training Size}
\label{app:retention_by_lang_ntrain}

\begin{figure*}[ht]
    \centering

    \begin{subfigure}[t]{0.48\textwidth}
        \centering
        \includegraphics[width=\linewidth]{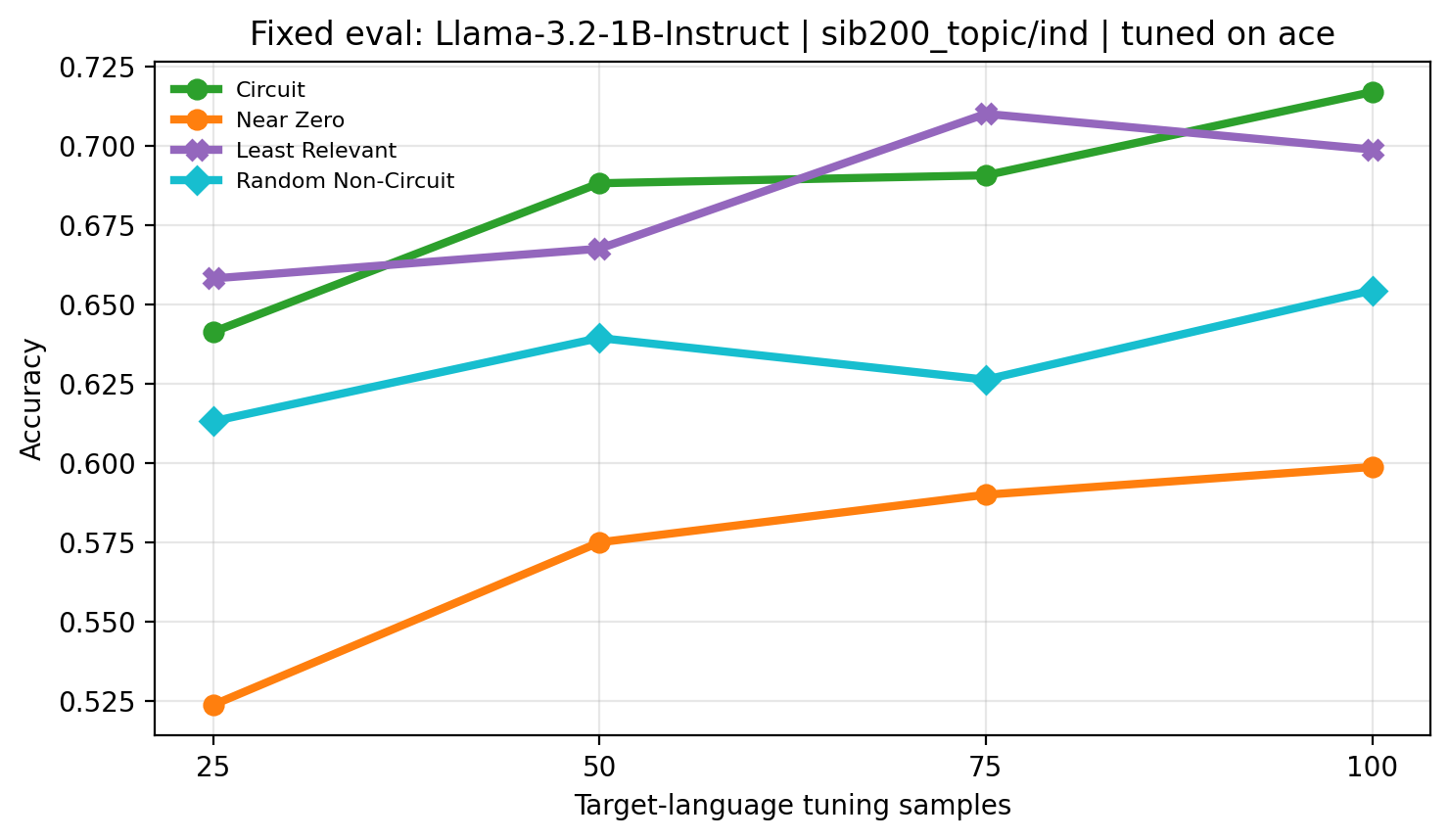}
        \caption{Tuned on Acehnese}
    \end{subfigure}
    \hfill
    \begin{subfigure}[t]{0.48\textwidth}
        \centering
        \includegraphics[width=\linewidth]{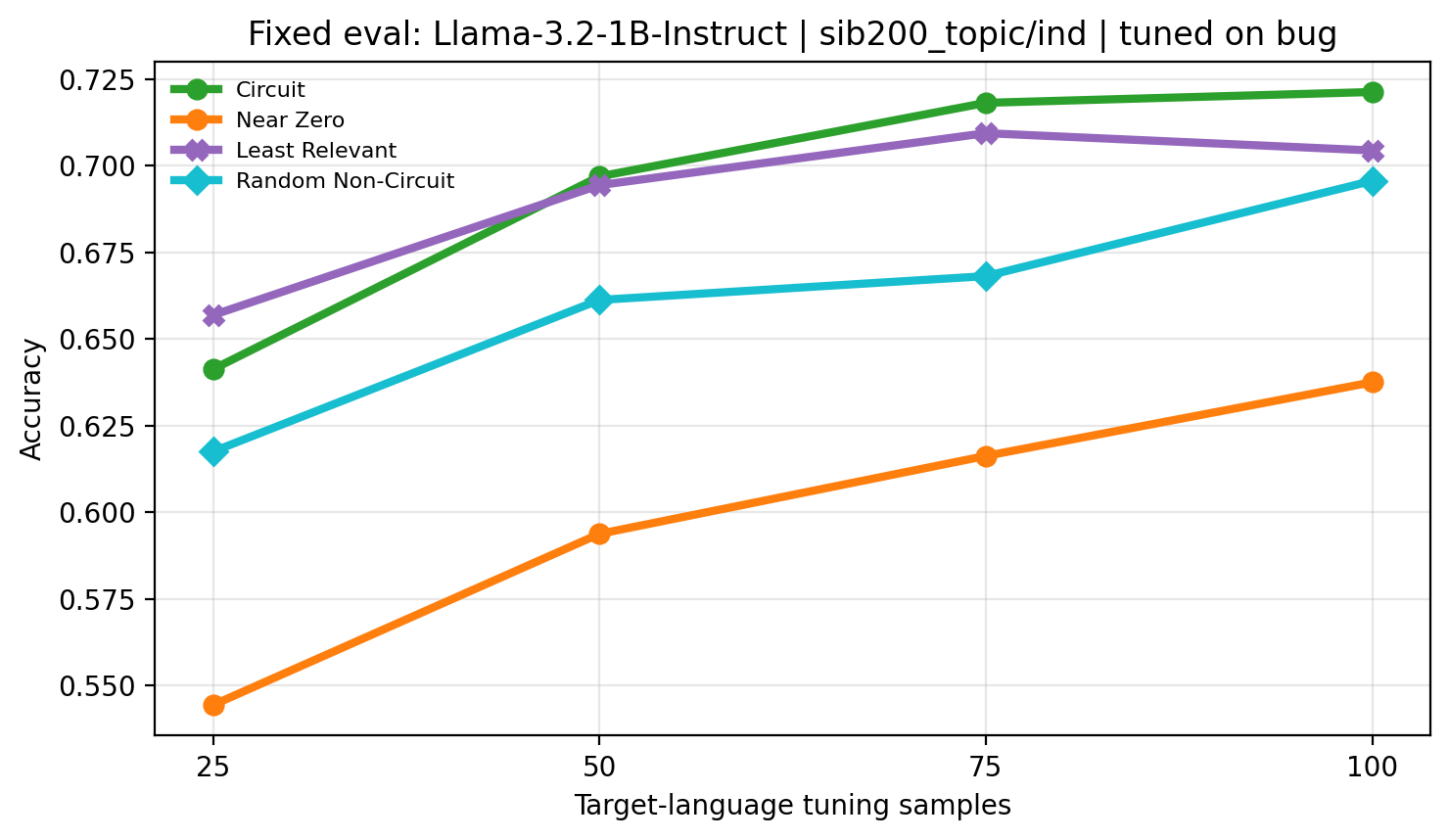}
        \caption{Tuned on Buginese}
    \end{subfigure}

    \vspace{0.8em}

    \begin{subfigure}[t]{0.48\textwidth}
        \centering
        \includegraphics[width=\linewidth]{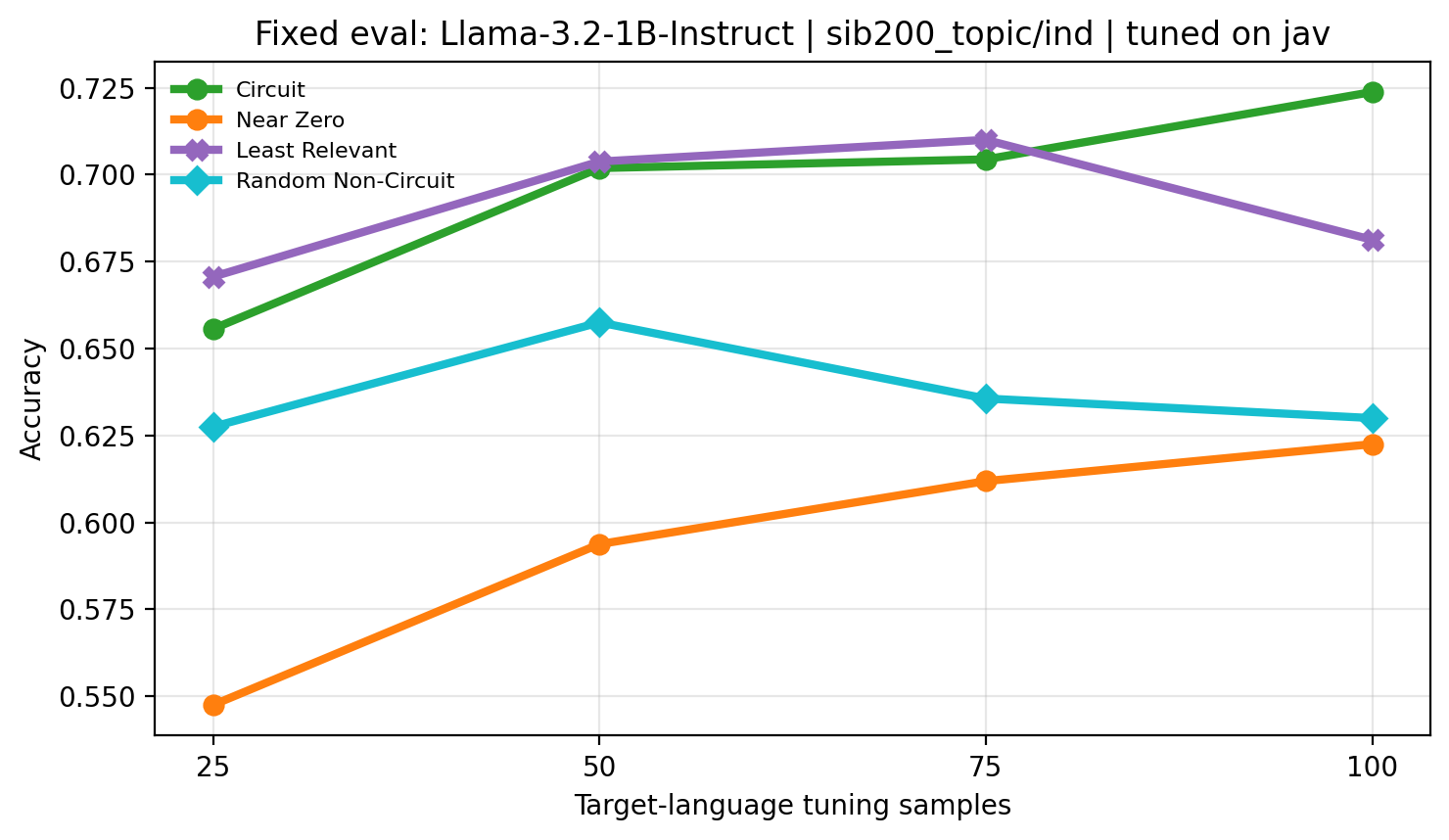}
        \caption{Tuned on Javanese}
    \end{subfigure}
    \hfill
    \begin{subfigure}[t]{0.48\textwidth}
        \centering
        \includegraphics[width=\linewidth]{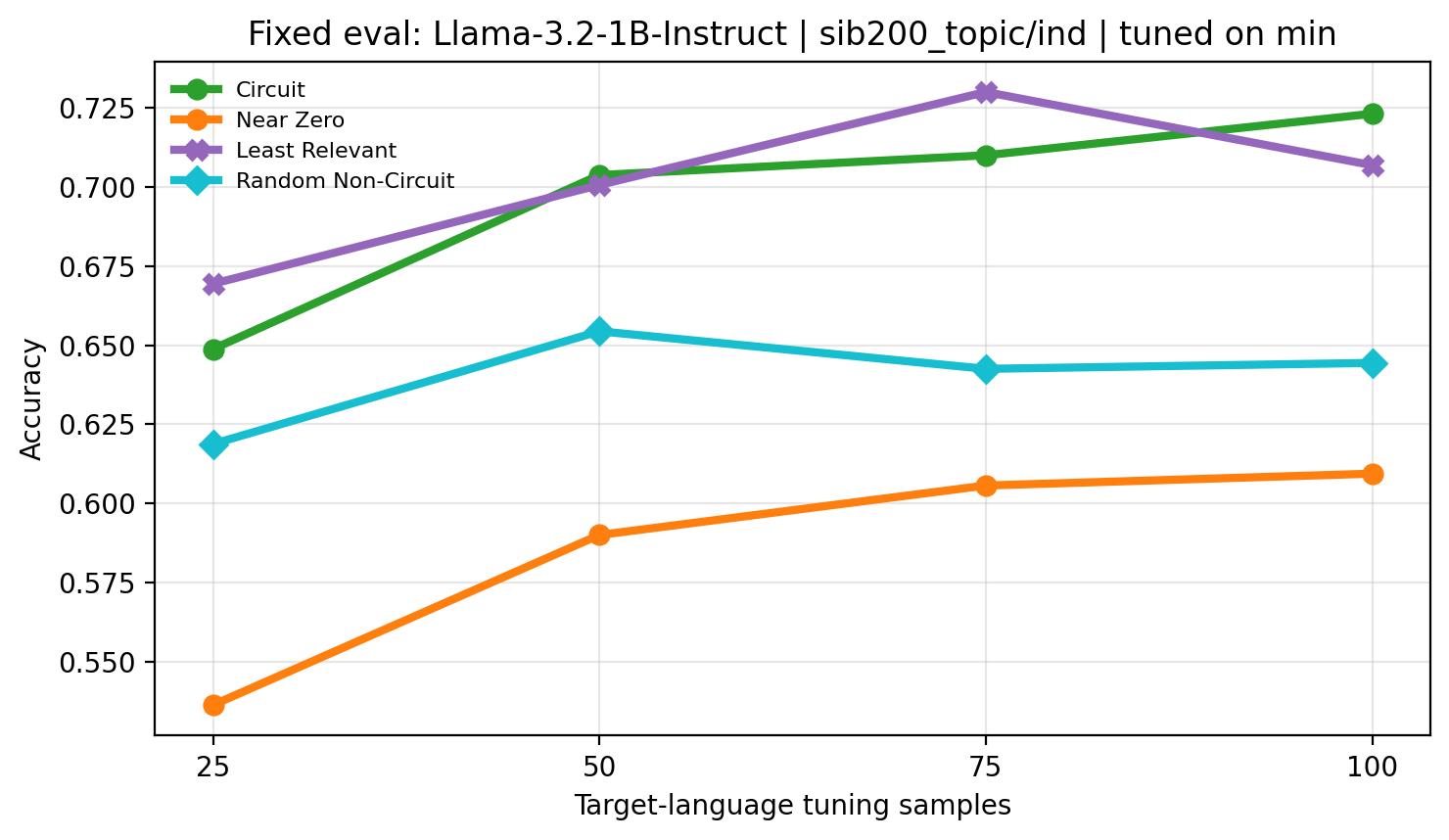}
        \caption{Tuned on Minangkabau}
    \end{subfigure}

    \caption{
    SIB200 topic retention by target-language training size.
    Each panel shows Indonesian SIB200 topic accuracy after adapting the model to a single target language.
    Indonesian tuning results are omitted.
    }
    \label{fig:sib200_by_lang_ntrain}
\end{figure*}

\begin{figure*}[ht]
    \centering

    \begin{subfigure}[t]{0.48\textwidth}
        \centering
        \includegraphics[width=\linewidth]{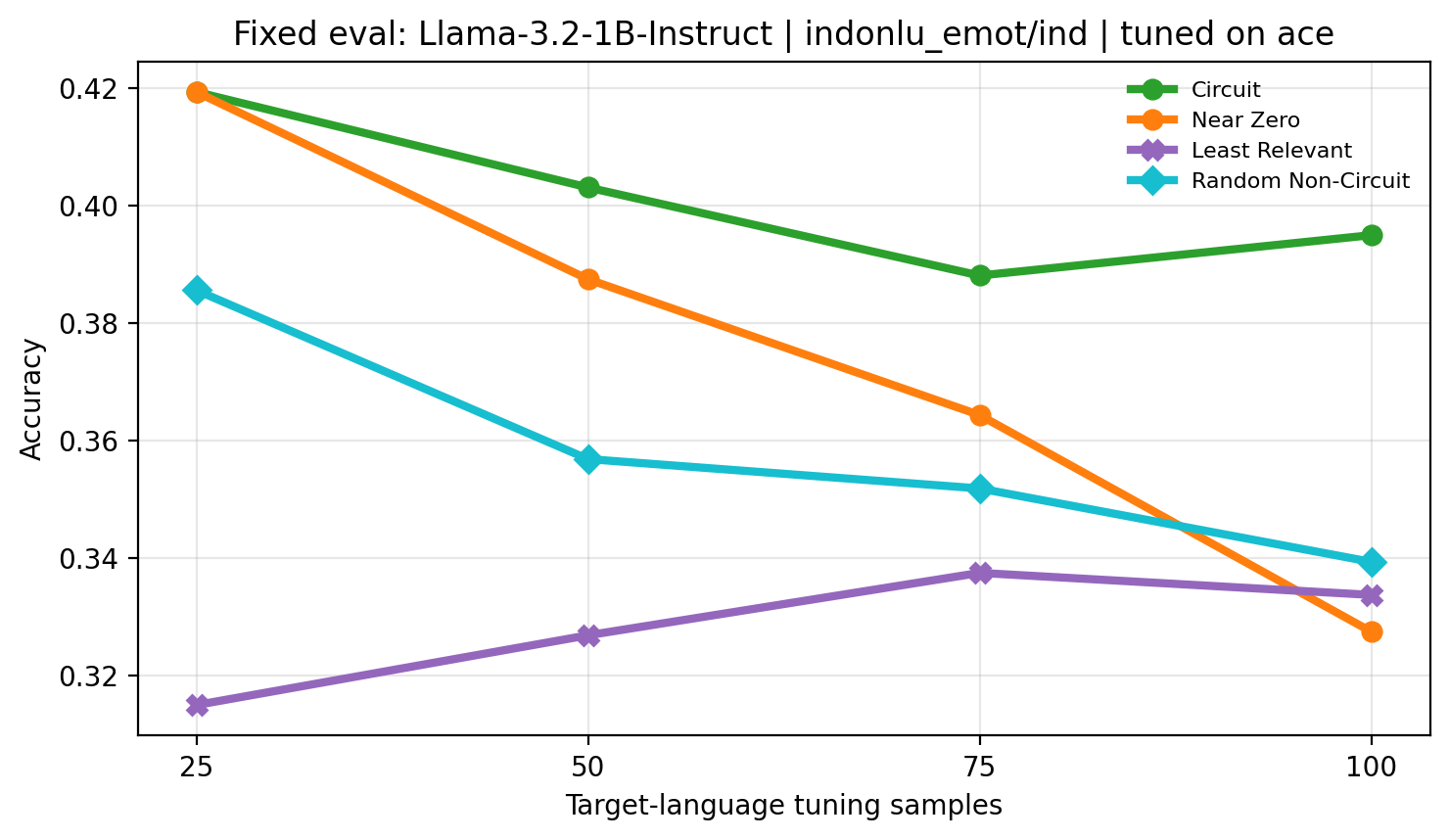}
        \caption{Tuned on Acehnese}
    \end{subfigure}
    \hfill
    \begin{subfigure}[t]{0.48\textwidth}
        \centering
        \includegraphics[width=\linewidth]{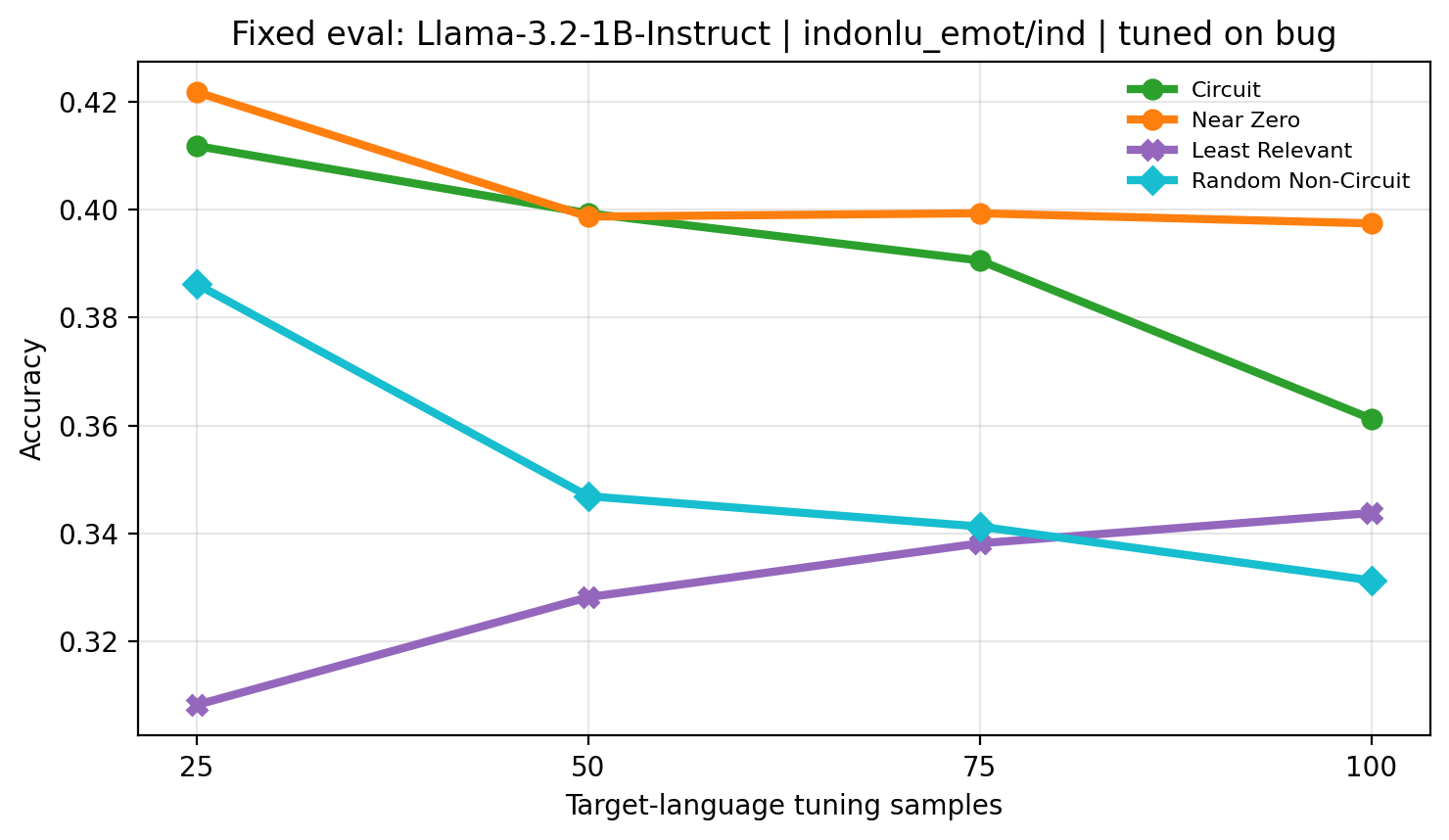}
        \caption{Tuned on Buginese}
    \end{subfigure}

    \vspace{0.8em}

    \begin{subfigure}[t]{0.48\textwidth}
        \centering
        \includegraphics[width=\linewidth]{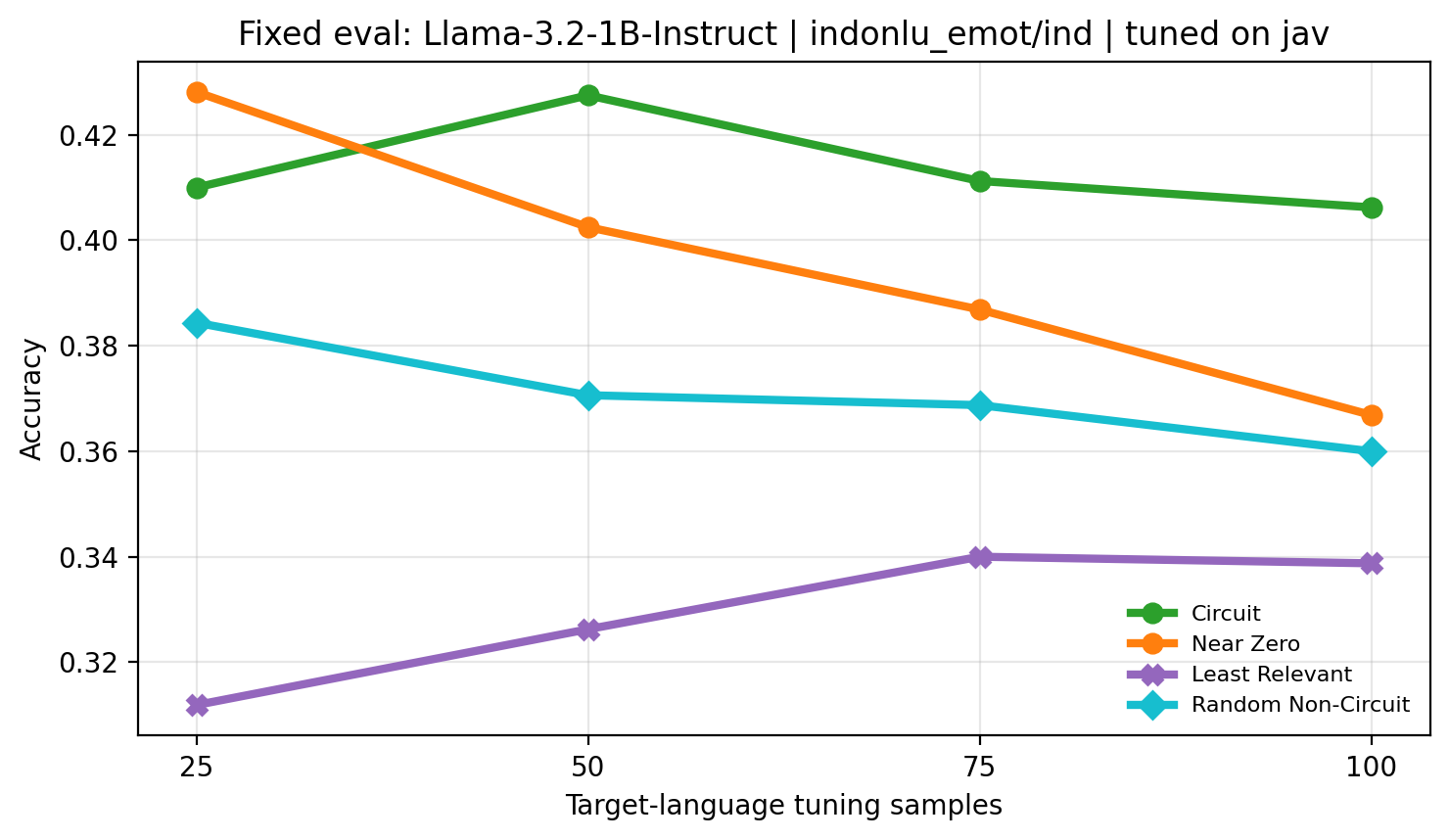}
        \caption{Tuned on Javanese}
    \end{subfigure}
    \hfill
    \begin{subfigure}[t]{0.48\textwidth}
        \centering
        \includegraphics[width=\linewidth]{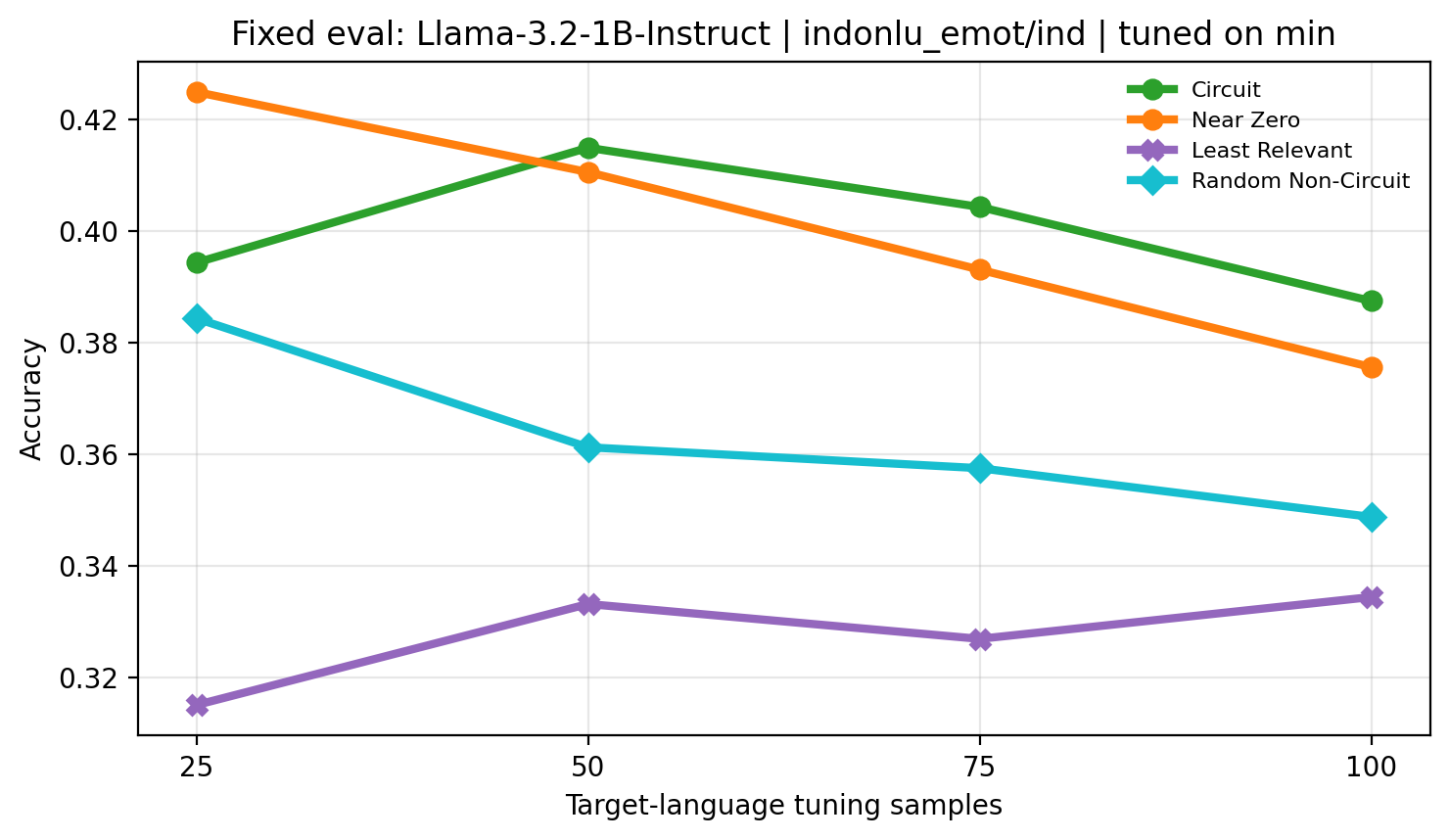}
        \caption{Tuned on Minangkabau}
    \end{subfigure}

    \caption{
    IndoNLU EmoT retention by target-language training size.
    Each panel shows Indonesian emotion-detection accuracy after adapting the model to one target language.
    Indonesian tuning results are omitted.
    }
    \label{fig:emot_by_lang_ntrain}
\end{figure*}

Figures~\ref{fig:sib200_by_lang_ntrain} and~\ref{fig:emot_by_lang_ntrain} show fixed-task retention across target-language training sizes for the main Llama-1B NusaX setting. 
These plots complement the aggregate preservation results in Table~\ref{tab:transfer_preservation_by_lang_llama1b} by showing whether retention behavior is stable across $n \in \{25,50,75,100\}$, rather than being driven by a single target-data budget.

For each target language, the model is adapted to that target language and then evaluated on fixed Indonesian retention tasks. 
SIB200 Topic retention tests broader, less-related classification behavior, while IndoNLU EmoT tests affective behavior that is closer to sentiment classification. 
We omit Indonesian tuning results because the main transfer setting adapts from Indonesian to local target languages, and the goal here is to measure how target-language adaptation affects fixed Indonesian behavior.

The trends show that retention behavior differs substantially across evaluation axes. 
Topic retention is the stricter preservation test: methods that are competitive for target transfer can still diverge strongly in their ability to preserve less-related topic classification. 
Emotion retention is more sentiment-adjacent, so it is generally less diagnostic of broad behavioral preservation than SIB200 Topic. 
These plots therefore support the main text's argument that controlled adaptation should be evaluated beyond target-language accuracy alone.

\FloatBarrier

\clearpage
\onecolumn

\section{Additional Scalability Results}
\label{app:scalability_results}

This appendix presents additional scalability and robustness results beyond those in the main Llama-3.2-1B NusaX setting. We include two kinds of evidence. First, we report detailed Llama-3.2-3B-Instruct results to examine whether the controlled-adaptation patterns observed at 1B also appear at a larger model scale. Second, we report XNLI results for Qwen2.5-0.5B and Llama-3.2-1B to test whether the intervention and source-retention patterns extend beyond NusaX sentiment classification.

The Llama-3B results should be read as a scalability check rather than a complete retuning of all baselines. As in the main experiments, sparse controls are matched to the post-pruning Circuit head budget, and Full FT and LoRA are same-data/same-protocol reference baselines. The XNLI experiments are narrower: they evaluate cross-lingual transfer, English source retention, and intervention behavior after English competence tuning, but they do not replicate the full NusaX multi-axis preservation suite.

The results show that sparse methods remain competitive at 3B scale, but the preservation profile differs by update location.

\subsection{Llama-3B Sparse-Control Results}
\label{app:llama3b_sparse_controls}

Table~\ref{tab:llama3b_transfer_preservation_by_lang} reports the full per-language sparse-control comparison for Llama-3.2-3B-Instruct. 
This table complements the aggregate scalability results in Table~\ref{tab:llama3b_scalability_avg} by showing whether the average trends are consistent across Acehnese, Buginese, Javanese, and Minangkabau. 
The comparison includes target-language transfer, Indonesian source retention, IndoNLU Emotion retention, and SIB200 Topic retention.

The results show that sparse methods remain competitive at 3B scale, but the preservation profile differs by update location. 
Least-relevant gives the highest average transfer, and Random NC is competitive on source and Topic retention. 
However, Near-zero suffers a large Topic-retention drop, and Least-relevant also loses substantial Topic retention relative to Circuit. 
Thus, the 3B results support the main text's conclusion that sparse updates are not interchangeable: matching the update budget alone does not guarantee the same preservation behavior.

\begin{table*}[ht]
\centering
\footnotesize
\setlength{\tabcolsep}{3.5pt}
\renewcommand{\arraystretch}{1.08}
\begin{tabular*}{\textwidth}{@{\extracolsep{\fill}}llcccccc@{}}
\toprule
\textbf{Method} & \textbf{Evaluation}
& \textbf{Pre-adapt.}
& \textbf{Ace} & \textbf{Bug} & \textbf{Jav} & \textbf{Min} & \textbf{Avg.} \\
\midrule

Zero-shot
& Transfer
& --
& 29.00 & 19.00 & 36.00 & 36.50 & 30.12 \\

\midrule

Circuit 
& Transfer & --
& 55.98 & 44.03 & 68.31 & 66.38 & 58.68 \\
& Source & 56.00
& \best{77.27} & 76.02 & 77.31 & 77.36 & 76.99 \\
& Emotion & 34.50
& \best{36.34} & \best{35.33} & \best{38.09} & \best{37.58} & \best{36.84} \\
& Topic & 20.00
& 68.78 & 66.64 & 69.09 & 69.64 & 68.54 \\

\midrule

Least-relevant
& Transfer & --
& \best{\scoreDelta{57.19}{+1.20}} 
& \best{\scoreDelta{44.86}{+0.83}} 
& \scoreDelta{69.08}{+0.77} 
& \best{\scoreDelta{67.53}{+1.16}} 
& \best{\scoreDelta{59.66}{+0.99}} \\
& Source & 56.00
& \scoreDelta{76.86}{-0.41} 
& \best{\scoreDelta{76.16}{+0.14}} 
& \worst{\scoreDelta{76.70}{-0.61}} 
& \scoreDelta{77.08}{-0.28} 
& \scoreDelta{76.70}{-0.29} \\
& Emotion & 34.50
& \scoreDelta{34.94}{-1.41} 
& \scoreDelta{33.39}{-1.94} 
& \scoreDelta{35.80}{-2.30} 
& \scoreDelta{35.50}{-2.08} 
& \scoreDelta{34.91}{-1.93} \\
& Topic & 20.00
& \scoreDelta{57.09}{-11.69} 
& \scoreDelta{57.23}{-9.41} 
& \scoreDelta{58.45}{-10.64} 
& \scoreDelta{59.66}{-9.98} 
& \scoreDelta{58.11}{-10.43} \\

\midrule

Near-zero
& Transfer & --
& \worst{\scoreDelta{54.95}{-1.03}} 
& \scoreDelta{44.27}{+0.23} 
& \worst{\scoreDelta{66.78}{-1.53}} 
& \worst{\scoreDelta{66.05}{-0.33}} 
& \worst{\scoreDelta{58.01}{-0.66}} \\
& Source & 56.00
& \worst{\scoreDelta{76.09}{-1.17}} 
& \worst{\scoreDelta{75.61}{-0.41}} 
& \scoreDelta{76.72}{-0.59} 
& \worst{\scoreDelta{76.56}{-0.80}} 
& \worst{\scoreDelta{76.25}{-0.74}} \\
& Emotion & 34.50
& \worst{\scoreDelta{30.61}{-5.73}} 
& \worst{\scoreDelta{30.22}{-5.11}} 
& \worst{\scoreDelta{31.42}{-6.67}} 
& \worst{\scoreDelta{31.28}{-6.30}} 
& \worst{\scoreDelta{30.88}{-5.95}} \\
& Topic & 20.00
& \worst{\scoreDelta{45.70}{-23.08}} 
& \worst{\scoreDelta{43.59}{-23.05}} 
& \worst{\scoreDelta{45.97}{-23.12}} 
& \worst{\scoreDelta{46.11}{-23.53}} 
& \worst{\scoreDelta{45.34}{-23.20}} \\

\midrule

Random NC
& Transfer & --
& \scoreDelta{55.55}{-0.44} 
& \worst{\scoreDelta{42.72}{-1.31}} 
& \best{\scoreDelta{69.28}{+0.97}} 
& \scoreDelta{67.25}{+0.88} 
& \scoreDelta{58.70}{+0.02} \\
& Source & 56.00
& \scoreDelta{77.09}{-0.17} 
& \scoreDelta{76.14}{+0.12} 
& \best{\scoreDelta{77.38}{+0.06}} 
& \best{\scoreDelta{77.81}{+0.45}} 
& \best{\scoreDelta{77.11}{+0.12}} \\
& Emotion & 34.50
& \scoreDelta{33.12}{-3.22} 
& \scoreDelta{32.22}{-3.11} 
& \scoreDelta{34.31}{-3.78} 
& \scoreDelta{34.39}{-3.19} 
& \scoreDelta{33.51}{-3.32} \\
& Topic & 20.00
& \best{\scoreDelta{70.31}{+1.53}} 
& \best{\scoreDelta{69.39}{+2.75}} 
& \best{\scoreDelta{70.30}{+1.20}} 
& \best{\scoreDelta{71.19}{+1.55}} 
& \best{\scoreDelta{70.30}{+1.76}} \\

\bottomrule
\end{tabular*}
\vspace{-0.8em}
\caption{
Transfer and behavior preservation after target-language adaptation for Llama-3.2-3B-Instruct.
The Pre-adapt. column reports zero-shot reference performance before adaptation for behavior-preservation evaluations.
For Transfer, the zero-shot target-language accuracies are shown in the first row.
Adapted scores are percentages averaged across training sizes and seeds; parentheses show differences from Circuit.
Bold marks the best score per column, and gray shading marks the lowest score.
Random NC denotes random non-circuit updates.
Emotion denotes IndoNLU emotion retention, and Topic denotes SIB200 topic-classification retention.
}
\label{tab:llama3b_transfer_preservation_by_lang}
\end{table*}

\subsection{Llama-3B Reference Baselines}
\label{app:llama3b_reference_baselines}

Table~\ref{tab:app_llama3b_reference_baselines_by_lang} compares Circuit with Full FT and LoRA for Llama-3.2-3B-Instruct. 
These baselines are included as same-data and same-protocol references: they use the same target-language examples, training sizes, seeds, optimizer settings, prompt format, and evaluation pipeline as Circuit. 
They are not intended to represent exhaustively tuned method-specific optima.

The reference-baseline comparison shows a clear tradeoff. 
Full FT yields the highest average target transfer, but sharply degrades broader preservation metrics, especially IndoNLU Emotion and SIB200 Topic. 
LoRA preserves Emotion better than Circuit, but it substantially underperforms on Topic retention and is weaker on average transfer. 
Circuit therefore provides a more balanced scalability profile: it is not the best method on every metric, but it better preserves broader Indonesian classification behavior while remaining competitive for target-language adaptation.

\begin{table*}[ht]
\centering
\footnotesize
\setlength{\tabcolsep}{4pt}
\renewcommand{\arraystretch}{1.08}
\begin{tabular*}{\textwidth}{@{\extracolsep{\fill}}llccccc@{}}
\toprule
\textbf{Method} & \textbf{Evaluation}
& \textbf{Ace} & \textbf{Bug} & \textbf{Jav} & \textbf{Min} & \textbf{Avg.} \\
\midrule

Circuit 
& Transfer & 55.98 & \best{44.03} & \best{68.31} & 66.38 & 58.68 \\
& Source & \best{77.27} & \best{76.02} & \best{77.31} & 77.36 & \best{76.99} \\
& Emotion & 36.34 & 35.33 & 38.09 & 37.58 & 36.84 \\
& Topic & \best{68.78} & \best{66.64} & \best{69.09} & \best{69.64} & \best{68.54} \\

\midrule
Full FT
& Transfer 
& \best{\scoreDelta{60.20}{+4.22}} 
& \scoreDelta{43.66}{-0.38} 
& \scoreDelta{67.31}{-1.00} 
& \best{\scoreDelta{69.16}{+2.78}} 
& \best{\scoreDelta{60.08}{+1.41}} \\
& Source 
& \worst{\scoreDelta{72.14}{-5.12}} 
& \worst{\scoreDelta{57.41}{-18.61}} 
& \scoreDelta{76.36}{-0.95} 
& \best{\scoreDelta{79.91}{+2.55}} 
& \worst{\scoreDelta{71.45}{-5.54}} \\
& Emotion 
& \worst{\scoreDelta{0.67}{-35.67}} 
& \worst{\scoreDelta{0.56}{-34.77}} 
& \worst{\scoreDelta{1.98}{-36.11}} 
& \worst{\scoreDelta{0.83}{-36.75}} 
& \worst{\scoreDelta{1.01}{-35.82}} \\
& Topic 
& \worst{\scoreDelta{4.12}{-64.66}} 
& \worst{\scoreDelta{7.73}{-58.91}} 
& \worst{\scoreDelta{1.34}{-67.75}} 
& \worst{\scoreDelta{6.77}{-62.88}} 
& \worst{\scoreDelta{4.99}{-63.55}} \\

\midrule
LoRA
& Transfer 
& \worst{\scoreDelta{50.78}{-5.20}} 
& \worst{\scoreDelta{42.81}{-1.22}} 
& \worst{\scoreDelta{62.03}{-6.28}} 
& \worst{\scoreDelta{64.95}{-1.42}} 
& \worst{\scoreDelta{55.14}{-3.53}} \\
& Source 
& \scoreDelta{73.25}{-4.02} 
& \scoreDelta{72.69}{-3.33} 
& \worst{\scoreDelta{73.17}{-4.14}} 
& \worst{\scoreDelta{73.11}{-4.25}} 
& \scoreDelta{73.05}{-3.93} \\
& Emotion 
& \best{\scoreDelta{38.17}{+1.83}} 
& \best{\scoreDelta{37.69}{+2.36}} 
& \best{\scoreDelta{38.52}{+0.42}} 
& \best{\scoreDelta{38.28}{+0.70}} 
& \best{\scoreDelta{38.16}{+1.33}} \\
& Topic 
& \scoreDelta{32.48}{-36.30} 
& \scoreDelta{30.38}{-36.27} 
& \scoreDelta{33.92}{-35.17} 
& \scoreDelta{32.80}{-36.84} 
& \scoreDelta{32.39}{-36.14} \\

\bottomrule
\end{tabular*}
\vspace{-0.8em}
\caption{
Llama-3.2-3B-Instruct comparison against reference baselines after target-language adaptation.
Scores are percentages averaged across training sizes and seeds; parentheses show differences from Circuit.
Bold marks the best score per column, and gray shading marks the lowest score.
Emotion denotes IndoNLU emotion retention, and Topic denotes SIB200 topic-classification retention.
}
\label{tab:app_llama3b_reference_baselines_by_lang}
\end{table*}

\subsection{XNLI Intervention Robustness}
\label{app:xnli_intervention_robustness}

Table~\ref{tab:xnli_faithfulness_knockout_combined} reports XNLI intervention results for Qwen2.5-0.5B and Llama-3.2-1B. 
Unlike the NusaX experiments, XNLI requires an English competence-tuning stage before circuit discovery and target-language adaptation. 
We therefore use XNLI as a narrower robustness check for intervention behavior rather than as a full replication of the NusaX controlled-adaptation suite.

\begin{table*}[ht]
\centering
\footnotesize
\setlength{\tabcolsep}{3pt}
\renewcommand{\arraystretch}{1.05}
\begin{tabular}{
  >{\raggedright\arraybackslash}p{0.17\textwidth}
  >{\centering\arraybackslash}p{0.07\textwidth}
  >{\centering\arraybackslash}p{0.07\textwidth}
  >{\centering\arraybackslash}p{0.07\textwidth}
  >{\centering\arraybackslash}p{0.07\textwidth}
  >{\centering\arraybackslash}p{0.07\textwidth}
  >{\centering\arraybackslash}p{0.07\textwidth}
  >{\centering\arraybackslash}p{0.07\textwidth}
  >{\centering\arraybackslash}p{0.07\textwidth}
  >{\centering\arraybackslash}p{0.07\textwidth}
  >{\centering\arraybackslash}p{0.07\textwidth}
}
\toprule
\textbf{Method} 
& \multicolumn{5}{c}{\textbf{Faithfulness (\% full acc.)}} 
& \multicolumn{5}{c}{\textbf{Knockout (\% acc. drop)}} \\
\cmidrule(lr){2-6}
\cmidrule(lr){7-11}
& \textbf{En} & \textbf{Es} & \textbf{Th} & \textbf{Vi} & \textbf{Zh}
& \textbf{En} & \textbf{Es} & \textbf{Th} & \textbf{Vi} & \textbf{Zh} \\
\midrule

\multicolumn{11}{l}{\textit{Qwen2.5-0.5B, $n_{\mathrm{circuit}}=250$}} \\
\addlinespace[1pt]
Circuit 
& \textbf{52.48} & \textbf{57.14} & 67.04 & \textbf{57.68} & 52.19
& \textbf{43.74} & \textbf{40.04} & \textbf{29.91} & \textbf{42.68} & \textbf{40.45} \\

Least-relevant 
& 44.49 & 55.75 & 68.95 & 51.09 & \textbf{53.40}
& 9.60 & 1.22 & 9.70 & 9.64 & 14.73 \\

Near-zero 
& 45.21 & 56.84 & 67.17 & 51.52 & 50.24
& 19.87 & 11.34 & 8.02 & 13.63 & 21.15 \\

Random NC 
& 44.74 & 53.01 & \textbf{69.35} & 52.03 & 50.44
& 33.42 & 29.42 & 27.79 & 32.15 & 29.03 \\

\midrule
\multicolumn{11}{l}{\textit{Llama-3.2-1B, $n_{\mathrm{circuit}}=250$}} \\
\addlinespace[1pt]
Circuit 
& \textbf{78.00} & 67.99 & 69.07 & 67.64 & \textbf{69.21}
& \textbf{50.13} & \textbf{33.14} & 23.48 & \textbf{36.89} & \textbf{37.42} \\

Least-relevant 
& 71.24 & \textbf{73.61} & \textbf{70.51} & 71.20 & 66.55
& 22.52 & 28.72 & 25.80 & 21.84 & 27.24 \\

Near-zero 
& 72.01 & 72.03 & 66.37 & 71.36 & 68.35
& 12.68 & 23.76 & 18.21 & 18.76 & 31.26 \\

Random NC 
& 68.33 & 69.90 & 64.68 & \textbf{72.87} & 63.93
& 21.38 & 28.37 & \textbf{31.37} & 20.69 & 32.92 \\

\bottomrule
\end{tabular}
\vspace{-0.8em}
\caption{
XNLI intervention evaluation with $n_{\mathrm{circuit}}=250$. Scores are validation label-token accuracies averaged across four seeds. Faithfulness is the percentage of full-model accuracy retained when only selected heads are preserved; knockout is the relative accuracy drop after selected-head ablation. Random NC is a matched, non-circuit-sparse control.
}
\label{tab:xnli_faithfulness_knockout_combined}
\end{table*}

The intervention results show that Circuit generally has strong knockout effects across model families, especially compared with Least-relevant and Near-zero controls. 
This means that ablating the selected Circuit heads produces a larger relative accuracy drop than ablating most matched sparse-control heads. 
Faithfulness is more mixed, which is expected because keep-only interventions are stricter and can be sensitive to distributed redundancy. 
Together, the XNLI intervention results support the claim that the selected heads remain informative intervention targets beyond the main NusaX setting, while also showing that faithfulness and knockout capture different aspects of circuit behavior.

\subsection{XNLI Transfer and Source Retention}
\label{app:xnli_transfer_retention}

Tables~\ref{tab:qwen05b_xnli_robustness_delta} and~\ref{tab:llama1b_xnli_robustness_delta} report XNLI cross-lingual transfer and English source retention after target-language adaptation. 
The target languages are Thai, Vietnamese, Spanish, and Chinese, and all scores are averaged across training sizes and seeds. 
Parenthesized values show the signed difference from Circuit for the same target language and evaluation axis.

These results should be interpreted differently from the NusaX preservation results. 
For XNLI, the goal is to check whether circuit-guided sparse updates preserve English competence while adapting to new target languages. 
We do not evaluate the broader NusaX-style preservation suite here, because the English competence-tuning stage would make it difficult to separate the effects of competence tuning from the effects of CT-SFT. 
The XNLI tables therefore provide evidence about transfer and source retention, not about broad behavioral preservation across unrelated Indonesian tasks.

\begin{table*}[ht]
\centering
\footnotesize
\setlength{\tabcolsep}{2.5pt}
\renewcommand{\arraystretch}{1.05}
\begin{tabular}{llccccc}
\toprule
Method & Eval & th & vi & es & zh & Avg. \\
\midrule
\multicolumn{2}{l}{\textit{Comp.-tuned ckpt}} 
& 50.81 & 64.25 & 65.88 & 69.31 & 62.56 \\
\multicolumn{2}{l}{\textit{\quad(En retention ceiling: 74.38)}} 
& & & & & \\
\midrule

\multirow{2}{*}{Circuit}
  & Transfer    
  & 57.59 & 63.16 & 65.25 & 68.06 & 63.52 \\
  & Source (En) 
  & 70.25 & 71.81 & 72.14 & 73.77 & 71.99 \\

\addlinespace
\multirow{2}{*}{Least-relevant}
  & Transfer    
  & 59.05 (+1.46) & 63.31 (+0.15) & 66.89 (+1.64) & 68.14 (+0.08) & 64.35 (+0.83) \\
  & Source (En) 
  & 73.02 (+2.77) & \textbf{73.80} (+1.99) & 74.34 (+2.20) & 73.69 (-0.08) & \textbf{73.71} (+1.72) \\

\addlinespace
\multirow{2}{*}{Near-zero}
  & Transfer    
  & \textbf{59.30} (+1.71) & 63.41 (+0.25) & \textbf{67.28} (+2.03) & 68.39 (+0.33) & \textbf{64.59} (+1.07) \\
  & Source (En) 
  & \textbf{73.30} (+3.05) & 73.08 (+1.27) & 73.80 (+1.66) & \textbf{74.05} (+0.28) & 73.55 (+1.56) \\

\addlinespace
\multirow{2}{*}{Random NC}
  & Transfer    
  & 58.64 (+1.05) & \textbf{63.92} (+0.76) & 67.00 (+1.75) & \textbf{68.61} (+0.55) & 64.54 (+1.02) \\
  & Source (En) 
  & 70.14 (-0.11) & 72.30 (+0.49) & \textbf{74.39} (+2.25) & 73.58 (-0.19) & 72.60 (+0.61) \\

\addlinespace
\multirow{2}{*}{Full FT}
  & Transfer    
  & 36.70 (-20.89) & 38.09 (-25.07) & 42.70 (-22.55) & 40.44 (-27.62) & 39.48 (-24.04) \\
  & Source (En) 
  & 40.31 (-29.94) & 39.12 (-32.69) & 45.36 (-26.78) & 43.11 (-30.66) & 41.98 (-30.01) \\

\addlinespace
\multirow{2}{*}{LoRA}
  & Transfer    
  & 58.75 (+1.16) & 62.58 (-0.58) & 64.30 (-0.95) & 66.75 (-1.31) & 63.09 (-0.43) \\
  & Source (En) 
  & 69.50 (-0.75) & 70.77 (-1.04) & 73.27 (+1.13) & 73.00 (-0.77) & 71.63 (-0.36) \\

\bottomrule
\end{tabular}
\caption{
XNLI cross-lingual transfer and English retention for Qwen2.5-0.5B with 250-sample English competence tuning.
Scores are percentages averaged over four seeds and $n\in\{25,50,75,100\}$.
Parentheses show signed differences from Circuit for the same column.
\textbf{Bold} marks the best score per column among matched sparse methods.
}
\label{tab:qwen05b_xnli_robustness_delta}
\end{table*}

Table~\ref{tab:qwen05b_xnli_robustness_delta} shows the Qwen2.5-0.5B results. 
We next report the corresponding Llama-3.2-1B results under the same XNLI protocol. Overall, the XNLI results support the robustness of the intervention evidence while also showing that Circuit is not uniformly the highest-transfer sparse method. 
Matched sparse controls can sometimes slightly outperform Circuit on transfer or source retention, especially when the evaluation is limited to the original XNLI label space. 
However, Circuit remains competitive while being tied to source-task intervention evidence, which is the main distinction from unguided sparse controls.

\begin{table*}[t]
\centering
\small
\setlength{\tabcolsep}{2.5pt}
\renewcommand{\arraystretch}{1.05}
\begin{tabular}{llccccc}
\toprule
Method & Eval & th & vi & es & zh & Avg. \\
\midrule
\multicolumn{2}{l}{\textit{Comp.-tuned ckpt}} 
& 41.69 & 45.31 & 45.50 & 45.06 & 44.39 \\
\multicolumn{2}{l}{\textit{\quad(En retention ceiling: 47.00)}} 
& & & & & \\
\midrule

\multirow{2}{*}{Circuit}
  & Transfer    
  & \textbf{45.75} & \textbf{45.53} & 48.89 & 46.87 & 46.76 \\
  & Source (En) 
  & \textbf{45.67} & \textbf{48.33} & 47.19 & 46.97 & \textbf{47.04} \\

\addlinespace
\multirow{2}{*}{Least-relevant}
  & Transfer    
  & 44.47 (-1.28) & 44.95 (-0.58) & 50.27 (+1.38) & \textbf{47.14} (+0.27) & 46.71 (-0.05) \\
  & Source (En) 
  & 41.36 (-4.31) & 46.59 (-1.74) & 46.64 (-0.55) & 46.73 (-0.24) & 45.33 (-1.71) \\

\addlinespace
\multirow{2}{*}{Near-zero}
  & Transfer    
  & 44.17 (-1.58) & 44.48 (-1.05) & \textbf{50.41} (+1.52) & 46.92 (+0.05) & 46.50 (-0.26) \\
  & Source (En) 
  & 41.38 (-4.29) & 46.91 (-1.42) & 47.62 (+0.43) & 46.86 (-0.11) & 45.69 (-1.35) \\

\addlinespace
\multirow{2}{*}{Random NC}
  & Transfer    
  & 45.45 (-0.30) & 45.47 (-0.06) & 50.03 (+1.14) & 47.00 (+0.13) & \textbf{46.99} (+0.23) \\
  & Source (En) 
  & 43.03 (-2.64) & 47.53 (-0.80) & \textbf{48.00} (+0.81) & \textbf{47.17} (+0.20) & 46.43 (-0.61) \\

\addlinespace
\multirow{2}{*}{Full FT}
  & Transfer    
  & 33.42 (-12.33) & 34.12 (-11.41) & 33.91 (-14.98) & 34.73 (-12.14) & 34.05 (-12.71) \\
  & Source (En) 
  & 34.02 (-11.65) & 34.05 (-14.28) & 33.84 (-13.35) & 34.30 (-12.67) & 34.05 (-12.99) \\

\addlinespace
\multirow{2}{*}{LoRA}
  & Transfer    
  & 42.95 (-2.80) & 45.69 (+0.16) & 50.03 (+1.14) & 46.70 (-0.17) & 46.34 (-0.42) \\
  & Source (En) 
  & 43.61 (-2.06) & 47.92 (-0.41) & 47.28 (+0.09) & 46.92 (-0.05) & 46.43 (-0.61) \\

\bottomrule
\end{tabular}
\caption{
XNLI cross-lingual transfer and English retention for Llama-3.2-1B with 250-sample English competence tuning.
Scores are percentages averaged over four seeds and $n\in\{25,50,75,100\}$.
Parentheses show signed differences from Circuit for the same column.
\textbf{Bold} marks the best score per column among matched sparse methods.
}
\label{tab:llama1b_xnli_robustness_delta}
\end{table*}

\end{document}